\documentclass[a4paper,11pt]{article}

\PassOptionsToPackage{numbers, compress}{natbib}
\usepackage{Definitions}

\usepackage{fullpage}
\usepackage{courier}
\usepackage{authblk}
\usepackage{calc}

\usepackage[vlined,ruled]{algorithm2e}
\usepackage{algorithmic}
\usepackage[utf8]{inputenc} % allow utf-8 input
\usepackage[T1]{fontenc}    % use 8-bit T1 fonts
\usepackage{hyperref}       % hyperlinks
\usepackage{url}            % simple URL typesetting
\usepackage{booktabs}       % professional-quality tables
\usepackage{amsfonts}       % blackboard math symbols
\usepackage{nicefrac}       % compact symbols for 1/2, etc.
\usepackage{microtype}      % microtypography
\usepackage{graphicx}
\usepackage{subfig}
\usepackage{xcolor}
\usepackage{paralist}
\usepackage{mathtools}
\usepackage{enumitem}

\newcommand*{\defeq}{\coloneqq}

\newcommand*{\history}[1]{\ensuremath{\Hcal_{#1}}}

\newcommand*{\actions}[1]{\ensuremath{\Acal_{#1}}}
\newcommand*{\feedbacks}[1]{\ensuremath{\Fcal_{#1}}}

\newcommand*{\historyAgent}[1]{\ensuremath{{\actions{}}_{#1}}}
\newcommand*{\historyFeedback}[1]{\ensuremath{{\feedbacks{}}_{#1}}}

\newcommand*{\actionCtx}{\ensuremath{\Ycal}}

\newcommand*{\policyTheta}{\ensuremath{p^*_{\actions{}; \theta}}}
\newcommand*{\policyThetaB}{\ensuremath{\mathbb{P}^*_{\actions{}; \theta}}}

\newcommand*{\uStar}{\ensuremath{\lambda^{*}_{\theta}}}
\newcommand*{\vStar}{\ensuremath{m^{*}_{\theta}}}

\newcommand*{\feedback}{\ensuremath{p^*_{\feedbacks{}; \phi}}}

\newcommand*{\uFeedback}{\ensuremath{\lambda_{\phi}^{*}}}
\newcommand*{\vFeedback}{\ensuremath{m_{\phi}^{*}}}

\newcommand*{\reward}{\ensuremath{R^{*}}}
\newcommand*{\grad}{\ensuremath{\nabla}}
\newcommand*{\prob}{\ensuremath{\PP{}}}

\newcommand*{\xhdr}[1]{\vspace{1mm}\noindent{{\bf #1.}}}

\newcommand*{\redqueen}{\textsc{RedQueen}}
\newcommand*{\memorize}{\textsc{Memorize}}

\newcommand*{\st}{\ensuremath{\,|\,}}
\newcommand*{\given}{\ensuremath{\,|\,}}

\newtheorem*{problem*}{Problem definition}

\title{Deep Reinforcement Learning of\\Marked Temporal Point Processes}

\author{Utkarsh Upadhyay}
\author{Abir De}
\author{Manuel Gomez-Rodriguez}
\affil{%
  {Max Planck Institute for Software Systems \\ \{utkarshu, ade, manuelgr\}@mpi-sws.org}
}

\graphicspath{{./FIG/}{./}}

\begin{document}
% \nipsfinalcopy is no longer used

\date{}

\maketitle

\begin{abstract}
%  \begin{itemize}
%    \item We use Reinforcement Learning to control point processes.
%    \item Are able to optimize for several different domains, and learn policies for intuitive but difficult to optimize for loss functions.
%  \end{itemize}
%
In a wide variety of applications, humans interact with a complex environment by means of asynchronous stochastic 
discrete events in continuous time.
Can we design online interventions that will help humans achieve certain goals in such asynchronous setting?
In this paper, we address the above problem from the perspective of deep reinforcement learning of marked temporal point 
processes, where both the actions taken by an \emph{agent} and the feedback it receives from the \emph{environment} are 
asynchronous stochastic discrete events characterized using marked temporal point processes. 
In doing so, we define the agent'{}s policy using the intensity and mark distribution of the corresponding process and then derive 
a flexible policy gradient method, which embeds the agent'{}s actions and the feedback it receives into real-valued vectors using 
deep recurrent neural networks. 
Our method does not make any assumptions on the functional form of the intensity and mark distribution of the feedback and it allows 
for arbitrarily complex reward functions.
We apply our methodology to two different applications in personalized teaching and viral marketing and, using data gathered
from Duolingo and Twitter, we show that it may be able to find interventions to help learners and marketers achieve their goals 
more effectively than alternatives.
\end{abstract}

% \vspace*{-6mm}
\section{Introduction}
\label{sec:introduction}
In recent years, the framework of marked temporal point processes (MTPPs)~\cite{Aalen2008} has become increasingly popular for modeling
asynchronous event data in continuous time,
which is ubiquitous in a wide range of application domains, from social and information networks to finance or health informatics.
For example,
in social and information networks, events may represent users'{} posts, clicks or likes;
in finance, they may represent buying and selling orders; or,
in health informatics, they may represent when a patient exhibits different symptoms or receives treatment.
In most cases, the development of a new model reduces to the problem of designing an appropriate functional form for the conditional intensity (or intensities)
of the events of interest as well as the distribution of the corresponding mark(s).

%  In recent years, marked temporal point processes (MTPPs) have emerged as powerful tools for modeling sequence of events localized in time, and therefore are used in
%  a wide variety of applications \eg, social activity modeling~\cite{redqueen17wsdm,control18jmlr,Farajtabar2014}, improving human learning~\cite{tabibian2017optimizing},
% product adoption~\cite{Valera2014},
%  reducing spread of misinformation~\cite{kim2018leveraging}, opinion diffusion~\cite{de2016learning}, \etc{}~In each of these applications, the underlying events have
% different meanings. For example,
%  in online social activity modeling, the events indicate posting of messages, whereas in human learning platforms, they denote the review of an item.
% However, in all these systems, the actual functional form governing the generation of events are complex, unknown and stochastic.

In this context, a recent line of work~\cite{kim2018leveraging,tabibian2017optimizing,wang2018stochastic,wang2017variational,control18jmlr,redqueen17wsdm} has
exploited an alternative view of MTPPs as stochastic differential equations (SDEs) with jumps~\cite{hanson2007applied} to design online, adaptive interventions using
stochastic optimal control.
While this line of work has shown promise at enhancing the functioning of social and information systems, their wide spread use and deployment is
precluded mainly by two drawbacks.
First, they make strong assumptions about the functional form of the conditional intensities and mark distributions of the MTPPs, which in turn prevent
them from using state of the art MTPP models based on deep learning~\cite{du2016recurrent, jing2017neural, MeiE16}.
Second, the objective functions that the interventions optimize upon, need to be carefully chosen to ensure that the underlying stochastic optimal
control problem remains tractable. As a consequence, the use of (more) meaningful objective functions with clear semantics is often off limits.
In our work, we overcome these drawbacks by approaching the problem from the perspective of deep reinforcement learning
of MTPPs.

% there has been an increasing interest in the area of control of MTPPs from the perspective of stochastic optimal control, where the goal is to steer the dynamics of the MTPP
% by means of interventions to maximize a certain objective function.
% However, they make strong assumptions about the functional form of the underlying dynamics which in turn constrains the expressive power of the corresponding model.
% While Kim~\ea{}~\cite{kim2018leveraging} simultaneously estimate the model parameters and control the process, they still assume a restrictive parametric form of the
% underlying dynamics.
%  Another problem which most of the previous works suffer from is that the choice of the objective function has to made very carefully lest the underlying optimal control
% problem should become intractable~\cite{control18jmlr,redqueen17wsdm,kim2018leveraging}.
% Furthermore, in these works, the objective functions are carefully chosen to make the underlying optimal control problem tractable and analytically solvable.
% This precludes the use of a wide range of other intuitive objective functions.

More specifically, we first introduce a novel reinforcement learning problem where both the actions taken by an \emph{agent} and the feedback it receives from
its \emph{environment} are asynchronous stochastic events in continuous time, which are characterized using MTPPs. Here, the goal is finding the \emph{optimal} intensity
and mark distribution for the agent'{}s actions---the optimal policy---that maximize an arbitrary reward function, which may depend on its actions and the feedback.
Then, we derive a novel policy gradient method, specially designed to solve the above problem, which embeds the agent'{}s actions and the feedback from
the environment into real-valued vectors using deep recurrent neural networks (RNNs).
%
 % We design a novel deep reinforcement learning  framework for control of point processes, which ameliorates the above limitations.
%
% Our proposed  framework consists of two components: (i) an agent, and (ii) an environment, where the agent observes the environment, and
% takes actions to steer the dynamics of the environment to maximize an arbitrary reward function.
%
% More in detail, we model the actions by the agent, and the observations from the environment, using MTPP which characterizes the continuous time interval
% between consecutive events (actions/observations), as a conditional intensity function.
%
% Then, we embed the actions and observations into real-valued vectors using deep recurrent neural networks, to capture the general structure of the underlying dynamics.
%
% Finally, we rigorously derive a novel policy gradient method to solve the control problem.
%
%  compute the optimal intensity function for the actions.
%
In contrast with the literature on stochastic optimal control of SDEs with jumps, our method does not make any assumptions on the functional form of the conditional intensity (or intensities) and mark distribution(s)
characterizing the feedback, and it allows for arbitrarily complex reward functions.
Moreover, it departs from previous work in the reinforcement learning literature~\cite{doya2000reinforcement,duan2016benchmarking,farajtabar2017fake,fremaux2013reinforcement,lillicrap2015continuous,mnih2016asynchronous,sutton1998reinforcement,vasilaki2009spike,wierstra2007solving} in two key aspects,
which are also illustrated in Figure~\ref{fig:rl-setting}:
\begin{figure}[t]
    \centering
    \includegraphics[width=0.7\textwidth]{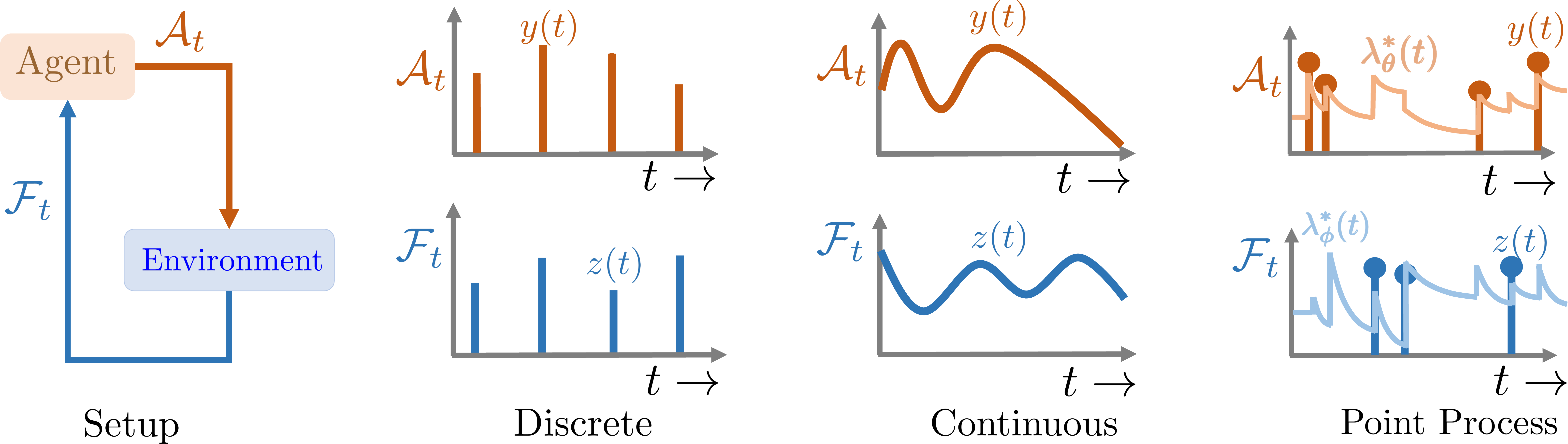}
    \caption{Reinforcement learning setups. In the traditional discrete time setting~\cite{sutton1998reinforcement}, actions and feedback occur in discrete time;
    in the continuous time setting~\cite{doya2000reinforcement}, actions and feedback are real value functions in continuous time;
    and, in the marked temporal point process setting (our work), actions and feedback are asynchronous events localized in continuous time.}\label{fig:rl-setting}
\end{figure}
\begin{itemize}[noitemsep,nolistsep]
\item[I.] The agent'{}s actions and environment'{}s feedback are asynchronous stochastic events in con\-ti\-nuous time.
In contrast, previous work has considered synchronous actions and (potentially delayed) feedback in discrete time~\cite{duan2016benchmarking,lillicrap2015continuous,mnih2016asynchronous,wierstra2007solving},
with few notable exceptions~\cite{doya2000reinforcement,fremaux2013reinforcement,vasilaki2009spike}. While these exceptions considered continuous time, they assumed actions and feedback to be continuous and
deterministic and the dynamics of the environment to be known.\footnote{\scriptsize Our setting should not be confused with the \emph{asynchronous} setting of Mnih~\ea{}~\cite{mnih2016asynchronous}, where the gradient descent is asynchronous but the action/observations are synchronous and the system evolves at discrete time steps.}

\item[II.] Our policy is a conditional intensity function (and a mark distribution), which is used to \emph{sample} the times (and marks) of the agent'{}s actions. Here, note that a sampled agent'{}s action may need to
be resampled due to the occurrence of new feedback events before the sampled time.
In contrast, previous works considered the policy to be a probability distribution or, more rarely, a deterministic function~\cite{doya2000reinforcement,fremaux2013reinforcement,vasilaki2009spike}.
\end{itemize}
%
% manuel: somewhere we need to perhaps justify why discretizing time is suboptimal
%
Finally, we apply our methodology to two different applications in  personalized teaching~\cite{leitner1972so,reddy2016unbounded,tabibian2017optimizing}
and viral marketing~\cite{karimi2016smart,spasojevic2015post,wang2018stochastic,control18jmlr,redqueen17wsdm}, respectively.
For \emph{simple} dynamics and objective fun\-ctions, which allow for stochastic optimal control approaches, our method achieves a comparable performance even though it does not have access to the true underlying dynamics.
For \emph{complex} dynamics and/or objective functions, which do not allow for stochastic optimal control approaches, our method is able to successfully find interventions that optimize the corresponding objective function and
beat several competitive baselines.
To facilitate research in temporal point processes within the reinforcement learning community at large, we are releasing an open-source implementation of our method in TensorFlow
as well as synthetic and real-world data used in our experiments.\footnote{\scriptsize\url{https://github.com/Networks-Learning/tpprl}}

\section{Problem formulation}
\label{sec:formulation}
% \vspace{-3mm}
In this section, we first briefly revisit the theoretical framework of marked temporal point processes~\cite{Aalen2008} and then use it to formally define our novel reinforcement
learning problem, where an agent interacts with a complex environment by means of asynchronous stochastic discrete events in continuous time.

\xhdr{Marked temporal point processes}
A marked temporal point process (MTPP) is a random process whose realization consists of an ordered sequence of events localized in time, \ie,
\begin{equation*}
\Hcal = \{ e_0 = (t_0, z_0), e_1 = (t_1, z_1), \dots, e_n = (t_n, z_n) \},
\end{equation*}
where $t_i \in \RR^{+}$ is the time of occurrence of event $i \in \ZZ$ and $z_i \in \Zcal$ is the associated mark.
The actual meaning of the events varies across applications, \eg\, in social networks, $t_i$ may represent the time when a message is posted, clicked
or liked, $z_i$ may represent the type of interaction, the message content, or its polarity,
%
% in case of posting of messages on social media sites~\cite{redqueen17wsdm,control18jmlr}, $t_i$ and $z_i$ represent the time and the content
% of the post respectively; in case of online learning platforms for review\footnote{For example,
% one word in a foreign vocabulary may be considered an item and a \emph{review} may ask the user to recall the item and
% show the user the correct answer after the attempt. This is the model followed by several flashcard based platforms, \eg, Anki, Duolingo.}
% by learners\footnote{\red{Cite memorize here?}}, the mark $z_i$  denotes the item selected and $t_i$ indiactes the time of review.
%
and the domain of the marks $\Zcal$ is application dependent.
Here, we characterize the event times of a MTPP using a conditional intensity function $\lambda^*(t)$, which is the probability of observing an event in the time window
$[t, t+dt)$ given the events history $\history{t} = \{ e_i = (t_i, z_i) \in \Hcal \st t_i < t \}$, \ie,
\begin{equation}
  \lambda^*(t) := \prob\{\text{event in } [t, t+dt) \given \history{t} \},
  \label{eqn:intensity}
\end{equation}
where the sign ${}^*$ means that the intensity may depend on the history $\history{t}$.
Moreover, we characterize the marks of the events using a distribution $m(z \given \history{t}) = m^*(z)$, which is the probability that mark $z$ is selected, \emph{if} an event
has occurred at time $t$.
Then, we can compute the likelihood of a history of events $\historyAgent{T} \subseteq \history{T}$ as:
\begin{equation}
  \PP(\historyAgent{T}) := \left( \prod_{e_i\in \historyAgent{T}} \overbrace{\lambda^{*}(t_i)}^{\text{Prob. of an action at } t_i}\, \underbrace{m^{*}(z_i)}_{\text{Prob. of mark } z_i} \right) \overbrace{\exp \left( - \int_{0}^{T} \lambda^{*}(s)\,ds \right)}^{\text{Prob. of no actions at } t\, \in\, [0, T] \setminus \{ t_i \}}.
\end{equation}
In the remainder of the paper, whenever an intensity function and mark distribution are parametrized by $\theta$, we write $\uStar(\cdot)$, $\vStar(\cdot)$, $\PP_{\theta}(\historyAgent{T})$, and,
for notational simplicity, use $p^{*}_{\theta} = (\uStar, \vStar)$ as a short-hand to denote the joint probability density of the MTPP.
Recent literature~\cite{du2016recurrent,farajtabar2017fake,karimi2016smart,kim2018leveraging,MeiE16,wang2017variational,control18jmlr} has established that MTPPs outperform other models (\eg, exponential law) in their ability to accurately predict online and off-line human actions.

% manuel: perhaps not necessary the line below?
% with $(t, x) \sim p_{\theta}$ denote sampling an event from the MTPP.
%
% An MTPP is completely characterized by the intensity $\uStar(\cdot)$ and the mark dis $\vStar(\cdot)$)
%
% It is related to the conditional density function $f_t^*(t)$ as~\cite{Aalen2008}:
% \begin{equation}
%   f_t^*(t) = \lambda^*(t) \exp{\left(-\int_{t_{n}}^{t} \lambda^*(s) ds\right)}.
%   \label{eqn:survival}
% \end{equation}
%

\xhdr{Reinforcement learning of marked temporal point processes}
Assume there is an agent who takes actions in a complex environment and the environment also provides feedback to the agent over
time.
Moreover, both the actions and the feedback are asynchronous stochastic events localized in time and thus we characterize them using
marked temporal point processes (MTPPs), \ie,
\begin{itemize}[noitemsep,nolistsep]
\item[---] \emph{Action events}: $\actions{} = \{ e_i = (t_i, y_i) \}$, where $(t_i, y_i) \sim \policyTheta = (\uStar, \vStar)$
\item[---] \emph{Feedback events}: $\feedbacks{} = \{ f_i = (t_i, z_i) \}$, where $(t_i, z_i) \sim \feedback = (\uFeedback, \vFeedback)$
\end{itemize}
In the above characterization, we allow the joint probability densities $\policyTheta$ and $\feedback$ to depend on the joint
history of events $\history{t} := \Acal_t \cup \Fcal_t$.
Finally, after a \emph{cut-off} time $T$, we assume that the agent receives an arbitrary (stochastic) reward $\reward(T)$, which may depend
on the agent'{}s actions $\actions{T}$ and the environment'{}s feedback $\feedbacks{T}$.

Given the above problem setting, we can formally define our reinforcement learning (RL) problem for marked temporal point processes as
follows:
\begin{problem*}\label{defn:problem}
  % Given a system
  % \begin{equation*}
  %  \Scal = \left( \policyTheta = (\uStar, \vStar), \feedback = (\uFeedback, \vFeedback), \reward \right),
  % \end{equation*}
  %
  Given an agent with $\policyTheta = (\uStar, \vStar)$, an environment with $\feedback = (\uFeedback, \vFeedback)$ and an arbitrary stochastic reward $\reward(T)$, the goal
  is to find the optimal action intensity and mark distribution---the optimal policy---that maximize the expected reward. Formally,
  \begin{equation}
    \maxi_{\policyTheta(\cdot)} \quad \EE_{\historyAgent{T} \sim \policyTheta(\cdot), \historyFeedback{T} \sim \feedback(\cdot)} \left[ \reward(T) \right],
    \label{eq:opt-policy}
  \end{equation}
  where the expectation is taken over all possible realizations of the marked temporal point processes associated to the agent'{}s action events and the environment'{}s feedback
  events. In the remainder of the paper, we will denote the optimal policy using $\pi^{*}(\theta) = \argmax_{\policyTheta(\cdot)} \EE \left[\reward(T) \right]$.
  % manuel: line below is redundant, we already defined this before.
  %notation $\historyAgent{T} \sim \policyTheta(\cdot)$ means that each event $(t_i, y_i) \in \historyAgent{T}$ was sampled from $(t_i, y_i) \sim \policyNoStar(\cdot \given \history{t_i})$.
\end{problem*}

Note that the above definition departs from previous work on reinforcement learning~\cite{doya2000reinforcement,duan2016benchmarking,fremaux2013reinforcement,lillicrap2015continuous,mnih2016asynchronous,sutton1998reinforcement,vasilaki2009spike,wierstra2007solving} in several ways.
First, the agent'{}s actions and environment'{}s feedback are asynchronous stochastic events in con\-ti\-nuous time. Moreover, note that the agent may receive feedback from the environment asynchronously
at any time, not only after each of its actions. This is in contrast with previous work in the literature, which has only considered synchronous actions (and potentially delayed) feedback in discrete time (or, in some cases, continuous actions and feedback), as illustrated in Figure~\ref{fig:rl-setting}.
Second, our policy is defined by a conditional intensity function (and a mark distribution), which is used to \emph{sample} the times (and marks) of the agent'{}s actions. Here, note that a sampled agent'{}s action may need to
be resampled due to the occurrence of new feedback events before the sampled time. In contrast, previous work has used probability distributions (or, in some cases, deterministic functions) as policies.

Remarkably, the above problem definition naturally fits numerous problems in a wide variety of application domains, particularly in the context of social and information online systems.
%
% For example, in viral marketing in social networks, a user who aims to increase the visibility of her posts is the agent, the user takes an action when she posts a message, her followers'{} feeds is the environment and the
% visibility (or attention) she receives defines the reward.
%
For example, in personalized teaching in online learning platforms, the platform that shows content items to learners is the agent, the platform takes an action when it shows an item to a learner, the learners are the environment, and the
probability that the learner recalls an item defines the reward.
In viral marketing in social networks, a user who aims to increase the visibility of her posts is the agent, the user takes an action when she posts a message, her followers'{} feeds form the environment and the
visibility (or attention) she receives defines the reward.
In all these cases, the environment distribution $\feedback$ may be highly complex and thus our policy gradient method will only assume that it can sample from $\feedback$. In other words,
the environment distribution will be considered a \emph{black box}.

\section{Proposed policy gradient method}
\label{sec:control}
% \vspace{-3mm}
\begin{figure*}[t]
  \centering
  \subfloat[Data and representation]{\includegraphics[width=0.45\textwidth]{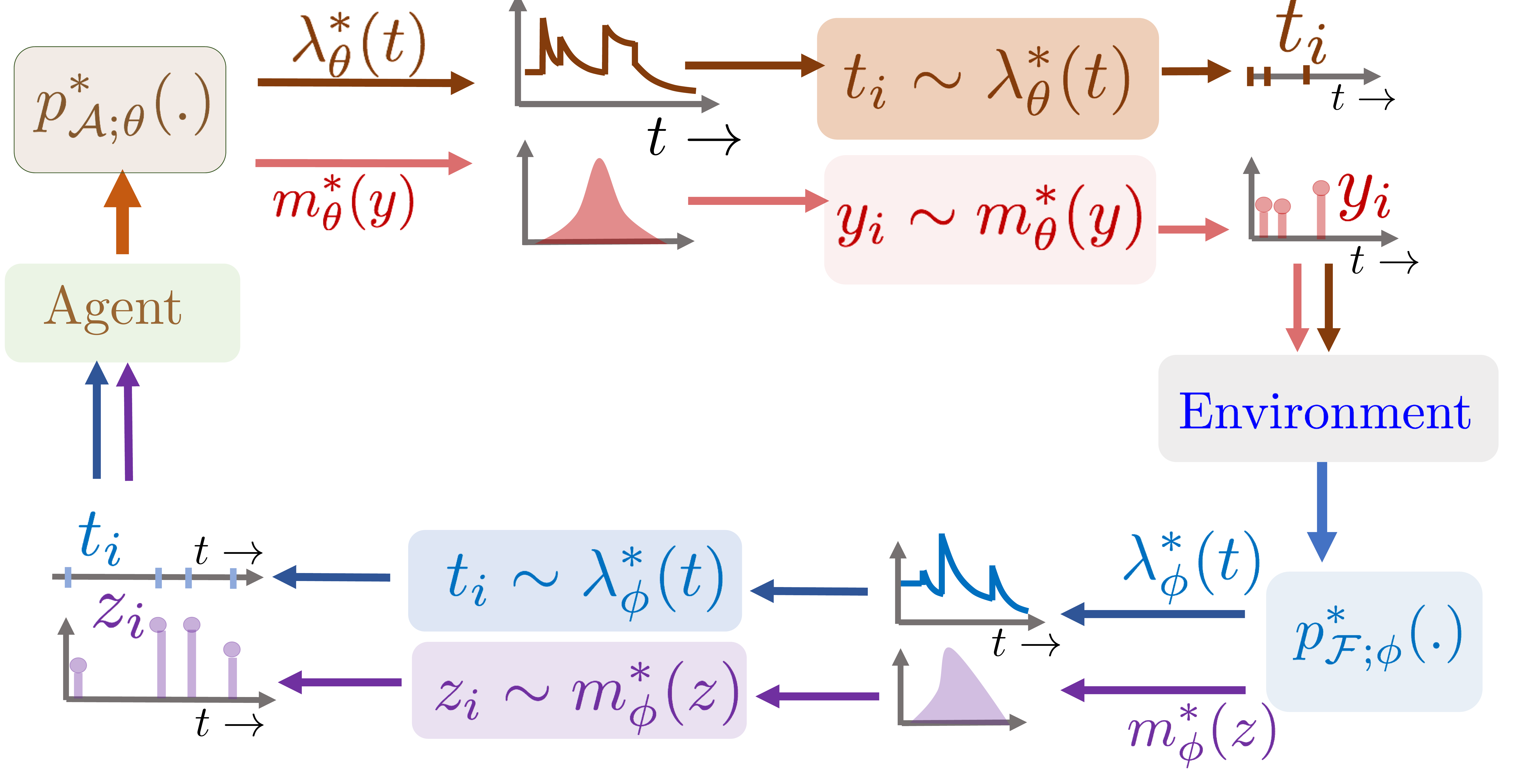}\label{fig:system}}\hspace{0.05\textwidth}%
  \subfloat[Policy parametrization]{\includegraphics[width=0.5\textwidth]{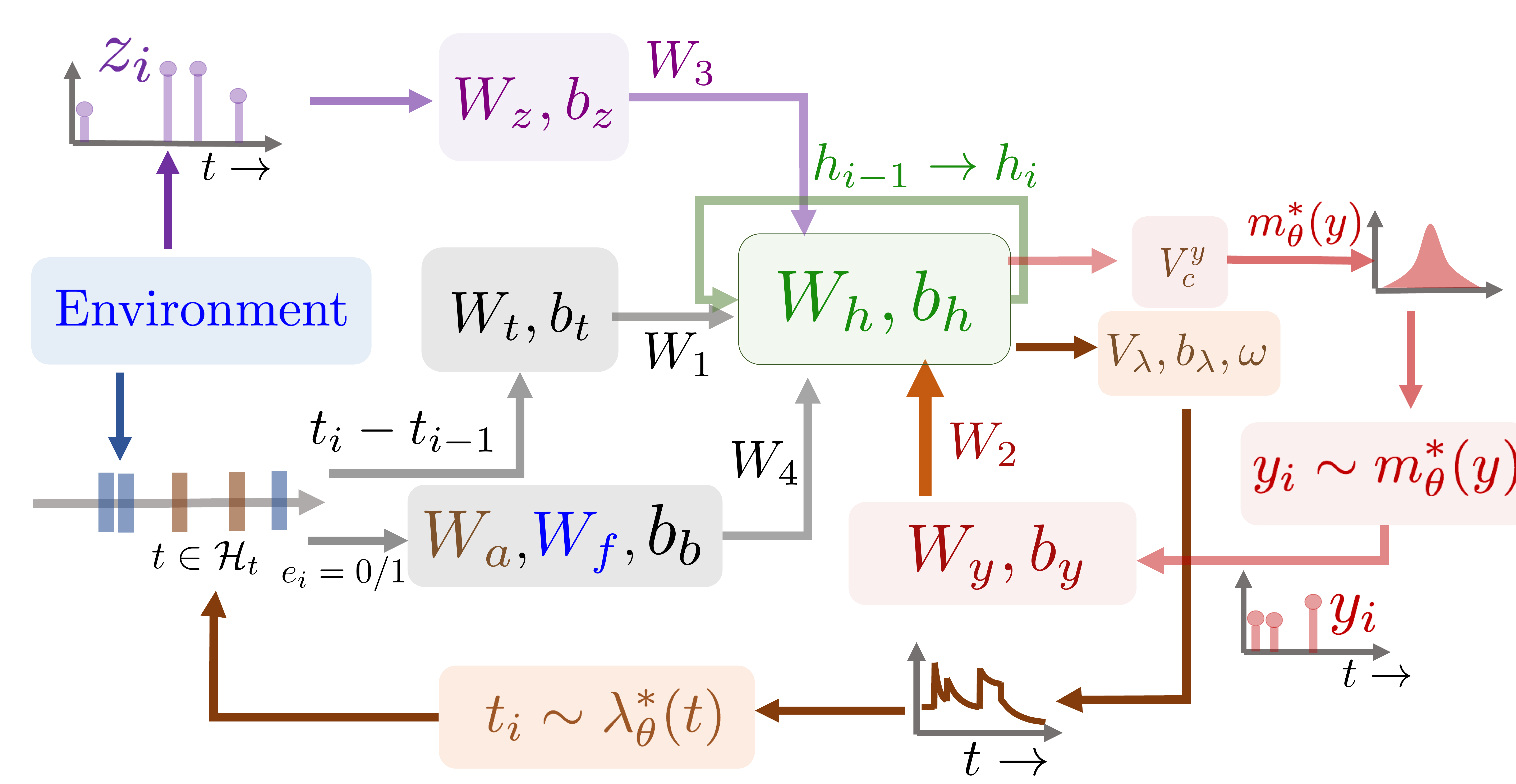}\label{fig:indep-arch}}%
  \caption{Reinforcement learning (RL) of of marked temporal point processes (MTPPs). Panel (a) shows the type of data and representation used in RL of MTPPs.
  Panel (b) shows the policy parametrization used by our policy gradient method.}\label{fig:arch}
  %
%   Panel (b) shows the neural-network architecture for our system.
  %
%  One of the inputs $y_i$ and $z_i$ will be marked as \emph{absent} using sentinel values depending on whether $b_i = 0$ or $b_i = 1$, respectively.
  %
\end{figure*}

In this section, we tackle the reinforcement learning problem defined by Eq.~\ref{eq:opt-policy} using a novel policy gradient method for marked temporal point processes.
More specifically, we first leverage recurrent neural networks (RNNs) to parametrize the policy $\policyTheta$ and then use stochastic gradient descent (SGD) to find the
policy parameters $\theta$ that maximizes the expected reward $\EE\left[\reward{}\right]$.

\xhdr{Policy parametrization}
In many application domains, at any time $t$, the (optimal) policy $\policyTheta$ that maximizes the reward may depend on the previous history of the action
events and the feedback events, $\history{t} = \actions{t} \cup \feedbacks{t}$, in an unknown and complex way.
To capture such dependence, we parametrize the policy $\policyTheta$ using a recurrent neural network (RNN), where we embed both the actions events and
the feedback events into real-valued vectors $\hb$, similarly as in several recent state of the art MTPP deep learning models~\cite{du2016recurrent, jing2017neural, MeiE16}\footnote{\scriptsize
Note that previous MTPP deep learning models aims to provide event predictions. This is contrast with the current work, which aims to provide optimal event interventions.}.
% we leverage the idea of Du \ea{}\cite{du2016recurrent} to capture such dependence using recurrent neural network parameterized by $\theta$.
%
Next, we elaborate further on our architecture\footnote{\scriptsize Depending on the application domains, action events or feedback events may not contain marks and, thus, the architecture may be
slightly simpler.}, which we also summarize in Figure~\ref{fig:arch}, and then discuss how to efficiently sample action events from the (optimal) policy.

% More specifically, the policy $\policyTheta$ at time $t$ acquires the following by time $t$we can express the policy  embed both the action events and feedback events into real valued vectors $\hb_i$
% ( Figure~\ref{fig:arch} ). In fact, $\hb_i$ captures the effect of history
% upto $i$-th event. Consequently, depending on the previous events, the policy $\policyTheta$ for the next action event $e_{i+1}\in \actions{T}$ can be represented as,
% \begin{align}
% \policyTheta(e_{i+1}|\history{t})=\policyTheta(e_{i+1}|\hb_i)
% \end{align}
% Such a formulation allows us to represent the history into the latent vector, find out the gradient of the reward (using Proposition~\ref{prop:gradient},\ref{prop-2}), and finally compute the optimal policy,
% without making any assumption of the underlying dynamics of the environment.

\noindent\hspace{0.1mm} --- \emph{Input layer.}
After the $i$-th event occurs, be it an action event or a feedback event, the input layer converts the associated information, \ie, the time $t_i$,
the marker $z_i$ (or $y_i$), and the type of event $e_i \in \{0, 1\}$, where $e_i=0$ denotes action and $e_i=1$ denotes feedback, into compact vectors.
Specifically, it computes:
\begin{align*}
\bm{\tau}_i &= \Wb_t (t_{i}-t_{i-1})+ \bb_t, &  \bm{y}_i = \Wb_y y_i+ \bb_y \,\,\, \mbox{if} \,\, e_i = 0 \\
\textbf{b}_i &= \Wb_{a} (1-e_i)+ \Wb_{f} e_i +\bb_{b}, & \bm{z}_i = \Wb_z z_i+ \bb_z \,\,\, \mbox{if} \,\, e_i = 1
\end{align*}
where $\Wb_{\bullet}$, $\bb_t$, $\bb_y$, $\bb_z$ and $\bb_b$ are trainable weights. Moreover, note that we encode the action marks $y_i$ and feedback marks $z_i$ separately since
they may belong to different domains. To this aim, one of the inputs $y_i$ and $z_i$ will be marked as \emph{absent} using sentinel values depending on whether $e_i = 0$ or $e_i = 1$,
respectively.
Finally, these signals are fed into the hidden layer, which we describe next.

\noindent\hspace{0.1mm} --- \emph{Hidden layer.} This layer iteratively updates the latent embedding $\hb_{i-1}$, by taking inputs of previous events from the input layer:
\begin{align}
 \hb_{i}=\text{tanh}(\Wb_h \hb_{i-1}+ \Wb_1 \bm{\tau}_i + \Wb_2 \bm{y}_i + \Wb_3 \bm{z}_i+\Wb_4 \textbf{b}_i +\bb_h),
\end{align}
where $\Wb_{\bullet}$ and $\bb_h$ are trainable weights.

\noindent\hspace{0.1mm} --- \emph{Output layer.} The output layer computes the policy $\policyTheta=(\uStar,\vStar)$, \ie, the intensity function $\uStar$ and the mark distribution $\vStar$. Assume the agent has generated
$i$ events by time $t$, then, the output layer computes the intensity as:
\begin{align}
  \uStar(t) =  \exp{\left( b_{\lambda} + w_t (t - t_i) + \Vb_{\lambda} \hb_i \right)}
  \label{eq:alt-intensity}
\end{align}
where $\Vb_{\lambda}$, $b_{\lambda}$ and $w_t$ are trainable weights and $t_i$ denotes the time of the $i$-th action event. Here, the $b_{\lambda}$ encodes
a base intensity level for the occurrence of the $(i+1)$-th action event, the term $w_t (t - t_i)$ encodes the influence of the $i$-th action event, and the term
$\Vb_\lambda$ encodes the influence of previous events.
%
% The intensity function represents both bursts ($ \Vb^T _h \hb_i\gg 0, w_t < 0$) and increasing urgency of generating events ($w_t > 0$). \etc{} \red{Correct signs for the sigmoidal intensity function.}
%
%
The particular choice of mark distribution $\vStar$ depends on the application domain. Here, we experiment with discrete marks and thus model the marks
using a multinomial distribution, \ie,
\begin{align} \label{eq:marks-distribution}
\prob[y_{i + 1} = c] = \frac{\exp(\Vb^{y}_{c, :} \hb_i )}{\sum_{l \in \Ycal}\exp(\Vb^{y}_{l, :} \hb_i)},
\end{align}
where $\Ycal$ denote the domain of the marks and $\Vb^{y}$ are trainable weights.
%
% \manuel{Saving space.. the sentence below is kind of trivial}
% For continuous marks, a sensible choice may be $y_{i + 1} \sim \Ncal\left( \mu_{\theta}(h_{i}), (\sigma_\theta(h_i) ^2 \right)$.
%
% Figure~\ref{fig:arch} summarizes shows the architecture of the neural network.

\xhdr{Sampling action events from the policy}
To implement the above policy $\policyTheta=(\uStar,\vStar)$, we need to be able to sample the action times $t$ and marks $y$
from the intensity function defined by Eq.~\ref{eq:alt-intensity} and the mark distribution defined by Eq.~\ref{eq:marks-distribution},
respectively.
While the latter reduces to sampling from a multinomial distribution, which is straightforward, the former requires developing a novel
sampling algorithm leveraging inverse transform sampling, which we describe in Algorithm~\ref{alg:sampling-brief}.
The details of calculating $CDF(\bullet)$ and the related modifications are provided in Appendix~\ref{app:sampling}.
\begin{algorithm}[t]                    % enter the algorithm environment
  \small
  \caption{Returns the next action time}
  \label{alg:sampling-brief}
  \begin{algorithmic}[1]
  \STATE \textbf{Input: } Parameters $b_\lambda, w_t, \Vb_\lambda, \hb_i$, last event time $t'$
  \STATE \textbf{Output: } Next action time $t$
    \STATE $CDF(\bullet) \leftarrow$ Cumulative distribution of next arrival time
    \STATE $u \leftarrow $\textsc{Unif}$[0, 1]$
    \STATE $t \leftarrow {CDF}^{-1}(u)$

    \WHILE{$t < T$}
      \STATE $(s, z)\leftarrow$\textsc{WaitUntilNextFeedback}$(t)$
      \IF{feedback arrived before $t$}
        \STATE $CDF(\bullet) \leftarrow $ \textsc{Modify}$(CDF(\bullet), s, z)$
        \STATE $t \leftarrow CDF^{-1}(u)$
      \ELSE
        \RETURN t
      \ENDIF
    \ENDWHILE
    \RETURN t
  \end{algorithmic}
\end{algorithm}

\xhdr{Maximizing the expected reward}
In the following, we denote the expected reward as a function of the policy parameters $\theta$ as:
\begin{align}
  J(\theta) &= \EE_{\historyAgent{T} \sim \policyTheta(\cdot), \historyFeedback{T} \sim \feedback(\cdot)} \left[ \reward(T) \right]
\label{eq:est-reward}
\end{align}
Then, we find the optimal policy $\policyTheta$ that maximizes the expected reward function $J(\theta)$ using stochastic gradient descent (SGD)~\cite{rumelhart1986learning}, \ie,
$\theta_{l + 1} = \theta_{l} + \alpha_{l} \grad_{\theta} J(\theta) |_{\theta = \theta_{l}}$.
% . That means,
% we iteratively update $\theta$ as follows:
%
%\begin{equation}
%  \theta_{l + 1} = \theta_{l} + \alpha_{l} \grad_{\theta} J(\theta) |_{\theta = \theta_{l}}.
%  \label{eq:theta-update}
%\end{equation}
%
To do so, we need to compute the gradient of the expected reward function $\grad_{\theta} J(\theta)$, however, this may seem challenging
at first especially since the expectation is taken over realizations of marked temporal point processes.
Perhaps surprisingly, we can compute such gradient using the following proposition (proved in Appendix~\ref{app:gradient}).
\begin{proposition}\label{prop:gradient}
   Given an agent with $\policyTheta = (\uStar, \vStar)$, an environment with $\feedback = (\uFeedback, \vFeedback)$, the gradient of the expected reward function $J(\theta)$ with
   respect to $\theta$ is given by:
  \begin{equation}
    \grad_{\theta} J(\theta) = \EE_{\historyAgent{T} \sim \policyTheta(\cdot), \historyFeedback{T} \sim \feedback(\cdot)} \left[ \reward(T) \grad_{\theta} \log{\PP_{\theta}(\historyAgent{T})} \right],
    \label{eq:grad}
  \end{equation}
  where $\log \PP_{\theta}(\historyAgent{T})= \sum_{e_i\in \actions{T}}\left(\log{\uStar(t_i)} + \log{\vStar(z_i)} \right) - \int_{0}^{T}\uStar(s)\,ds$.
%  \begin{equation}
%    \log \PP_{\theta}(\historyAgent{T})= \sum_{e_i\in \actions{T}}\left(\log{\uStar(t_i)} + \log{\vStar(z_i)} \right) - \int_{0}^{T}\uStar(s)\,ds
%  \label{eq:LL}
%  \end{equation}
\end{proposition}
In the above proposition, the gradient of the log-likelihood of the times and marks of a realization of the marked temporal point process associated to the agent'{}s actions,
$\grad_{\theta} \log{\policyThetaB(\history{T})}$, can be easily computed using the policy parametrization defined by Eqs.~\ref{eq:alt-intensity} and~\ref{eq:marks-distribution}.
Moreover, note that the proposition formally shows that the REINFORCE trick~\cite{williams1992simple} is still valid if the expectation is taken over realizations of marked temporal point processes, which are a type of \emph{random elements}~\cite{daley2007introduction} whose values are discrete events localized in continuous time.

Unfortunately, the above procedure does not limit the intensity of actions by the agent and this may be problematic in practice (\eg, in viral marketing in social networks, a user who aims to
increase the visibility of her posts may only be able to post a certain number of times).
To overcome this, we consider instead a penalized expected reward function $J_{r}(\theta)$ with differentiable regularizers $g_\lambda(\uStar(t))$ and $g_m(\vStar(t))$, which implicitly impose a budget on the number
of action events and marks, respectively, \ie,
\begin{align}
J_r(\theta)=\EE_{\historyAgent{T} \sim \policyTheta(\cdot), \historyFeedback{T} \sim \feedback(\cdot)} \left[ \reward(T) -q_l\int_0 ^T g_\lambda(\uStar(t)) -q_m\int_0 ^T g_m(\vStar(t))dt\right].
\end{align}
The gradient of the penalized reward can be readily computed using the following proposition (proved in Appendix~\ref{app:gradient-regularized}):
 \begin{proposition}\label{prop:gradient-regularized}
Given an agent with $\policyTheta = (\uStar, \vStar)$, an environment with $\feedback = (\uFeedback, \vFeedback)$, the gradient of $J_{r}(\theta)$ is given by,
  \begin{align}
  \grad_{\theta}J_r(\theta)
   &=  \EE_{\historyAgent{T} \sim  \policyTheta(\cdot), \historyFeedback{T} \sim \feedback(\cdot)} \Bigg[ \nonumber \\
    & \phantom{+} \left( \reward(T)- q_l \int_{0}^{T} g_\lambda(\uStar(t)) - q_m \int_{0}^{T} g_m(\vStar(t))\,dt \right) \grad_{\theta} \log{\PP_{\theta}(\historyAgent{T})} \nonumber\\
  & - \left. \left( q_l \int_{0}^{T} g'_\lambda(\uStar(t)) \grad_{\theta} \uStar(t)\,dt + q_m \int_{0}^{T} g'_m(\vStar(t)) \grad_{\theta} \vStar(t)\,dt \right) \right],
  \label{eq:reinforce-regularized}
\end{align}
where $g'_\lambda(\uStar(t)) = \frac{d\, g_\lambda(\uStar(t))}{d\, \uStar(t)}$ and $g'_m(\vStar(t)) = \frac{d\, g_m(\vStar(t))}{d\, \vStar(t)}$.
% , and $\PP_{\theta}(\historyAgent{T})$ is given by Eq.~\ref{eq:LL}.
\end{proposition}
%
%  \begin{equation*}
%    \log \policyThetaB(\history{T})= \sum_{e_i\in \actions{T}}\big(\log{\uStar(t_i)} + \log{\vStar(z_i)} \big) - \int_{0}^{T} \uStar(s)\,ds
%  \label{eq:LL}
%  \end{equation*}
%
%
In our experiments, we will approximate the expectation in Eq.~\eqref{eq:reinforce-regularized} by first running a \emph{batch} of realizations (or \emph{episodes}) of the corresponding
marked temporal point processes\footnote{\scriptsize In some applications, we may be able to play back historical data from the environment against our policy and, in other domains, we may need to
resort to a (complex) environment simulator.} and then calculating the mean of the resulting gradients for each batch.
%
%\manuel{I have removed the sentence below to save space and since it is not new}
%As this is a Monte Carlo estimation of the gradient $\grad_{\theta} J(\theta)$, it will converge to the true gradient with the rate $\Ocal(n^{-\frac{1}{2}})$, where $n$ is the batch size, under the assumption
%that the gradients have finite variance.

% \red{Should discuss these here?}
% \begin{itemize}
  % \item Differential reward, true continuous time.
%   \item Different forms of the intensity functions, compare with Mei~\ea{}~\cite{MeiE16}, and talk about probability of extinction.
  % \item Practical considerations?
%   \item \red{Tentative}: This reward, following conventions, can have two components: a differential reward accumulated over the trajectory $\int_{0}^{T} r(t)\,dt$ and a terminal reward, which depends on the final state of
% the system $\phi(\history{T})$. \red{These components can also be stochastic.} This formulation can allow us to reduce the variance in the gradient calculation \red{\ldots}~\cite{peters2008reinforcement}.
  % \item Generalizing to the case when we would like to provide more feedback to the process?
% \end{itemize}

% \vspace{-3mm}
% \section{Experiments}
% \label{sec:experiments}
% \vspace{-3mm}
% \begin{itemize}
%  \item We will test the performance of the policy gradient algorithms on two different problems from two different domains which can be cast as asynchronous control problems: the \emph{when-to-post} problem and the
% \emph{when-to-review} problem.
% \end{itemize}
\section{Experiments on spaced repetition} \label{sec:spaced-repetition}
%
% \manuel{Saving space}
% In this section, we evaluate our method in a personalized teaching application---spaced repetition~\cite{leitner1972so,reddy2016unbounded,tabibian2017optimizing}---and show
% that our method outperforms several alternatives.
%
%
\begin{figure}[t]
  \vspace{0.25cm}
  \centering
  \subfloat[Recall]{\includegraphics[width=0.3\textwidth]{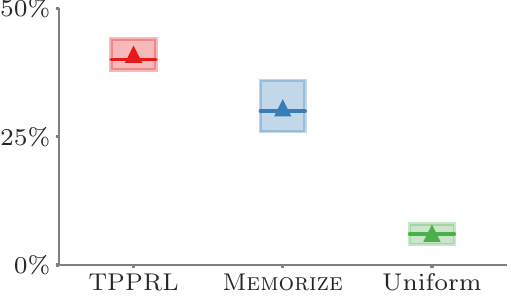}\label{fig:recall}}\hspace{0.04\textwidth}%
  \subfloat[Items' difficulty]{\includegraphics[width=0.3\textwidth]{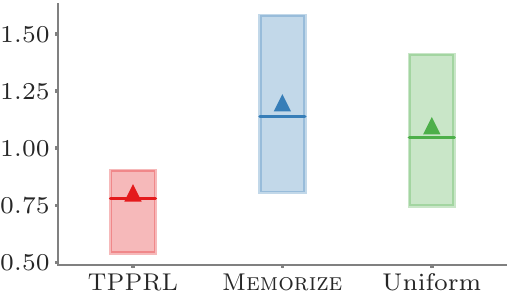}\label{fig:item-difficulty}}\hspace{0.04\textwidth}%
  % \subfloat[Example trace]{\includegraphics[width=0.3\textwidth]{example-755}\label{fig:trace}}
  \subfloat[Reviewing events]{\includegraphics[width=0.3\textwidth]{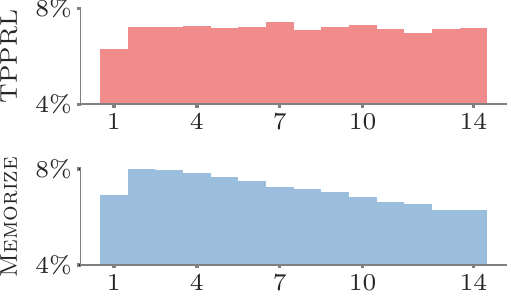}\label{fig:trace-spaced-repetition}}
  \caption{Spaced Repetition. Performance of our policy gradient method against \memorize{}~\cite{tabibian2017optimizing} and a uniform baseline, which follows a constant reviewing rate and
  chooses items uniformly at random.
  Panel (a) shows the empirical recall probability at time $T+\tau$ and Panel (b) shows the difficulty level of the items selected for review by different methods. In both cases, the solid horizontal line (triangle) shows the median (average) value across review sequences
 and the box limits correspond to the $25$\%-$75$\% percentiles.
 All methods schedule (within a small tolerance) the same number of review events.
  Panel (c) compares the average fraction of review events per day across all items for our method (above) and \memorize{} (below).
  }
  \label{fig:results-spaced-repetition}
\end{figure}

\xhdr{Problem definition}
It is well known in the psychology literature that repeated and temporally distributed reviewing of information aids long term memorization~\cite{leitner1972so, lindsey2014improving, metzler2009does, mettler2016comparison}. % (See~\cite{sr} for a detailed review).
Following recent work in the machine learning literature~\cite{mettler2016comparison, reddy2016unbounded, tabibian2017optimizing}, we will consider the following setting: an online learning platform
needs to teach one student some number of items with varying difficulty, say, words from the vocabulary of a foreign language. To this aim, the platform interacts with the student during a studying period
by asking her to \emph{review} each item multiple times, \ie, show a word to the student, ask for its translation, and then show the correct answer.
Then, the goal is to help the platform decide when to ask the student to review each item to better prepare her for a \emph{test}, which will take place sometime after
the learning period is over.
Under our problem definition,
the online platform is the agent,
it generates action events $\actions{}$ when it asks a student to review an item,
the student is the environment and she generates feedback events $\feedbacks{}$ when she reviews an item, indicating whether she was able to recall the item or not,
and the recall probability at the test time defines the reward.
%
%The agent's actions $\actions{}$, hence, are times it requests a student to review and the item it picks. The environment, \ie, the student, generates feedback events $\feedbacks{}$ at each review, indicating whether she was able
% to recall the item asked of her or not.
%

% Several heuristics have been proposed which offer to help a learner in determining when to review which item, starting from Leitner~\ea{}~\cite{} and followed by by several recent studies~\cite{,,pashler2009predicting, , }.
%
Interestingly, the above setting has been recently studied from the point of view of stochastic optimal control~\cite{tabibian2017optimizing}, where the authors have derived the optimal scheduling algorithm
for a set of items.
However, their solution assumes that the difficulty of the items and the student model are known~\cite{settles2016trainable} and that the objective function---the reward---has a particular functional form which depends
on the average recall probability over time (and not the actual sampled recall at test time).
%
% The same issues discussed above have plagued the solution they have proposed.
%
% Perhaps the most famous of these heuristics is the Leitner system
%
Here, we use our reinforcement learning method to derive (optimal) policies for arbitrarily complex and unknown student models, items with unknown difficulties and more intuitive reward definitions.
%
% As in previous work in psychology literature~\cite{pashler2009predicting,lindsey2014improving}, the objective of the platform will be to prepare the student for a \emph{test}, which will take place time $\tau$ after the learning period
% is over.
%
% \manuel{I moved the rest to experimental setup}

% objective function of how a student will perform on a test scheduled $\tau$ time after the learning session.

\xhdr{Experimental setup}
Since we cannot make real interventions in an online learning platform, we use data from Duolingo to fit a probabilistic student model, as reported in previous work~\cite{settles2016trainable, tabibian2017optimizing},
which we then use to simulate a student'{}s performance over time (refer to Appendix~\ref{app:student-model} for further details on the student model).
Here, the optimal policy $\policyTheta{} = (\uStar(t), \vStar(t))$ comprises of a reviewing intensity function and a multinomial mark distribution. The former characterizes
when to review and the latter characterizes which item to review each time.
Then, we train and test our policy gradient method as follows.

% Our agent's policy function $\policyTheta{}$ will consist of both a mark distribution $\vStar(t)$ as well as an intensity function $\uStar(t)$, denoting which item to review and when to review it, respectively.
%
% We will assume that have a simulator a (complex) simulator which can model the memory of a student.
%
Given a student model and a set of items, we train the platform'{}s policy $\policyTheta{}$ by using SGD with a quadratic (entropy) regularizer on the reviewing intensity (mark distribution),
\ie, $g(\uStar(t), \vStar(t)) = \left(\uStar(t)\right)^2 + H(\vStar(t))$ where $H(\vStar(t_i)) \defeq - \sum_{c \in \actionCtx} \PP[y_i = c] \log \PP[y_i = c]$, on a training consisting of simulated reviewing and test sequences.
More specifically, on iteration $i$, we build a batch of $b$ reviewing (or studying) sequences of time length $T$, where we sample student'{}s recalls from the student model every time our
policy $p^{*}_{\theta_i}$ generates a reviewing events and compute the reward at the end of each sequence.
Here, the reward is the sampled recall at test time $T + \tau$, which is a natural performance measure for the goal stated in the problem definition.
To test the trained model, we just generate additional reviewing sequences using the student model and the trained policy and compute the reward at the end of
each sequence.
Appendix~\ref{app:implementation} for further details on the training and testing procedure.
%
% The agent will schedule reviews for the student within the time period $[0, T]$ and the student will respond by telling the agent whether she was able to successfully recall the item or not.
%
% Here, we use whether the student recalls each item at a test time $T + \tau$ as reward.
%
Here, we compare the performance of our method with two alternatives: (i) a state of the art method called \memorize{}~\cite{tabibian2017optimizing} which, in contrast with our work, has full access to
the student model and is specially designed to maximize the average recall probability over time, and (ii) a baseline reviewing schedule which follows a constant reviewing rate and choose items uniformly
at random.

\xhdr{Results}
Figures~\ref{fig:results-spaced-repetition}(a-b) summarize the results, where the number of reviewing events by each method is the same. The results show that: (i) by maximizing the
actual reward one is aiming for, our method is able to outperform both \memorize{} and the baseline by large margins; and, (ii) given the limited study time, our method tends to focus on less
difficult items.
Finally, in Figure~\ref{fig:results-spaced-repetition}(c), we compare how our method and \memorize{} distribute reviewing events during the studying period. While our method keeps a constant
load over time, \memorize{} provides initially a heavier studying load.
\section{Experiments on smart broadcasting} \label{sec:smart-broadcasting}
%
% \manuel{Saving space}
% In this section, we evaluate our method in a viral marketing application, smart broadcasting~\cite{karimi2016smart,spasojevic2015post,wang2018stochastic,control18jmlr,redqueen17wsdm}, and
% show that our method outperforms several alternatives.

\xhdr{Problem definition}
In the smart broadcasting problem, first introduced by Spasojevic et al.~\cite{spasojevic2015post}, the goal is to help a social media user decide when to
post to achieve high visibility in her followers'{} feeds, \ie, to elicit attention from her followers.
Under our problem definition,
the user is the agent,
she generates action events $\actions{}$ when she posts,
her followers'{} feeds forms the environment,
the environment generates feedback events $\feedbacks{}$ when any of the other users her followers follow post,
and the visibility she receives defines the reward.
Then, the problem reduces to finding the (optimal) policy $\policyTheta$ that maximizes the reward.

Following previous work~\cite{wang2018stochastic,control18jmlr,redqueen17wsdm}, we measure visibility a user achieves, \ie, the reward, using
two different metrics:
(i) the position of her most recent post on her followers'{} feeds over time, or \emph{rank}, \ie, $\reward(T) = \int_{0}^{T} r(t) dt$, where the position zero, $r(t) = 0$, corresponds to top and
thus lower is better;
(ii) the (amount of) time that her most recent post is at the top of her followers'{} feeds, or \emph{time at the top}, \ie, $\reward(T) = \int_{0}^{T} \II(r(t)<1) dt$, and thus higher is better.
If the followers'{} feeds are sorted in reverse chronological order, previous work has derived optimal offline~\cite{karimi2016smart} and online~\cite{redqueen17wsdm}
algorithms for (i) and (ii), respectively, under the additional assumption that the posting intensity of other users her followers follow adopts certain functional form.
However, as pointed out by previous work, feeds are typically algorithmically sorted, the posting intensity of other users may be highly complex, and thus the derived
algorithms may be of limited use in practice.
Here, we use our reinforcement learning method to derive (optimal) policies for algorithmically sorted feeds and, by doing so, we are able to help users achieve higher
visibility than the above algorithms.
Appendix~\ref{app:feed-reverse-chronological-order} contains additional experiments for feeds sorted in reverse chronological order.
\begin{figure}[t]
  \centering
  % \subfloat[Average rank]{\includegraphics[width=0.25\textwidth]{algo-avg-rank}\label{fig:avg-rank}}\hspace{0.04\textwidth}%
  \subfloat[Average rank]{\includegraphics[width=0.3\textwidth]{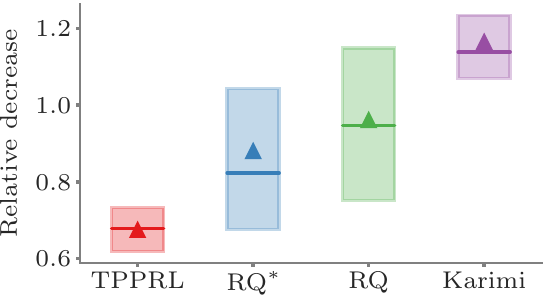}\label{fig:avg-rank}}\hspace{0.04\textwidth}%
  % \subfloat[Time at top]{\includegraphics[width=0.25\textwidth]{algo-top-1}\label{fig:top-1}}\hspace{0.04\textwidth}%
  \subfloat[Time at top]{\includegraphics[width=0.3\textwidth]{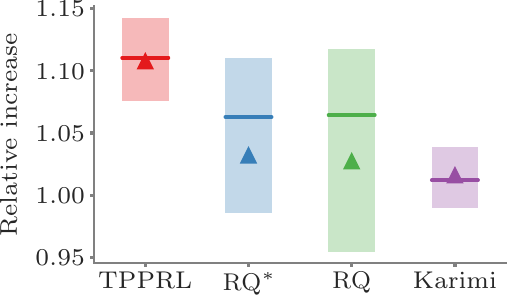}\label{fig:top-1}}\hspace{0.04\textwidth}%
  % \subfloat[Example trace]{\includegraphics[width=0.3\textwidth]{example-755}\label{fig:trace}}
  \subfloat[Example]{\includegraphics[width=0.3\textwidth]{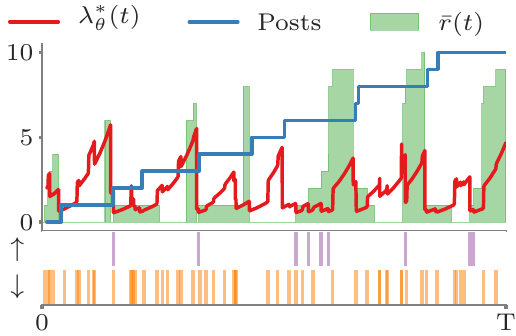}\label{fig:trace-smart-broadcasting}}
  \caption{Smart broadcasting. Performance of our policy gradient method against \redqueen{}~\cite{redqueen17wsdm} (RQ),
  a variant of \redqueen{} which has access to true ranks (RQ$^*$), and Karimi's method~\cite{karimi2016smart} on feeds using a sorting algorithm
  based on a priority queue (refer to Appendix~\ref{app:feed-sorting-algorithm}).
  Panels (a) and (b) show the average rank and time at the top, where the solid horizontal line shows the median value across users, normalized
  with respect to the value achieved by a user who follows a uniform Poisson intensity, and the box limits correspond to the $25$\%-$75$\% percentiles.
  For the average rank, lower is better and, for time at the top, higher is better. In both cases, the number of messages posted by each method is the
  same.
  Panel (c) shows a user'{}s intensity $\uStar(\cdot)$ (in blue), as provided by our method, the counts of the user'{}s posts (in green), the average
  rank (in red), the posting times of a competing user with higher priority (in purple), and the posting times of another competing user with lower
  priority (in yellow).
  }
  \label{fig:results-feed-sorting-algorithm}
\end{figure}

\xhdr{Experimental setup}
We use data gathered from Twitter as reported in previous work~\cite{cha2010measuring}, which comprises profiles of $52$ million users, $1.9$ billion directed follow links among
these users, and $1.7$ billion public tweets posted by the collected users.
The follow link information is based on a snapshot taken at the time of data collection, in September 2009.
%
% Here, we focus on the tweets published during a two month period, from July 1, 2009 to September 1, 2009, and sample $100$ users uniformly at random. For
% each of these users, we sample of one her followers along with five users this follower follows and reconstruct the follower'{}s (partial) feed by collecting all the (re)tweets published
% by these five users.
%
Here, we focus on the tweets published during a two month period, from July 1, 2009 to September 1, 2009, and sample $100$ users uniformly at random.
For each of these users, we retrieve five of her followers (chosen at random), select five other \emph{followees} of each follower (chosen at random), and collect all the (re)tweets they published.
Each follower represents a wall and our broadcaster is \emph{competing} with the other followees of follower for attention.

Since we do not have access to the feed sorting algorithm used by Twitter, we experiment with a relatively simple sorting algorithm based on a priority queue\footnote{\scriptsize We
expect that, the more complex the sorting algorithm, the larger the competitive advantage our algorithm will offer in comparison with competing methods designed for feeds sorted in
reverse chronological order.} (refer to Appendix~\ref{app:feed-sorting-algorithm}).
Here, since our feed sorting algorithm does only depends on the time of the post and the identity of the user who posts, not marks (\eg, content of the post), the optimal policy only comprises an intensity function, \ie, $\policyTheta = \uStar(t)$.
Then, we train and test our policy gradient method as follows.

For each user, we divide her feedback events, \ie, the posts by other users her followers follow, into a training set and a test set. The latter contains all feedback events generated in a time window
of length $T$ at the end of the recording period and the former contains all other feedback events.
Here, we set the length $T$ such that the overall expected number of events in the test set is $\sim$$200$.
Then, we train each user'{}s policy $\uStar(t)$ by using stochastic gradient descent (SGD) with a quadratic regularizer $g(\lambda^{*}(t)) = \left(\lambda^{*}(t)\right)^{2}$.
More specifically, on each iteration $i$, we build a batch of $b$ sequences of length $T$, taken uniformly at random from the training set, we \emph{replay} the feedback events from these
sequences while interleaving the posts generated by our policy $\lambda^{*}_{\theta_i}$, and compute the reward at the end of each sequence.
To test the trained policy $\uStar(t)$, we just replay the feedback events from the test set while interleaving the posts generated by the policy and compute the reward at the end
of the sequence.
Appendix~\ref{app:implementation} contain additional implementation details.

In the above, we experiment both with rank and time at the top as rewards and compare our method with two state of the art methods, \redqueen{}~\cite{redqueen17wsdm} and the method by Karimi et
al.~\cite{karimi2016smart}.
The former is an online algorithm specially designed to minimize the average rank in feeds sorted in reverse chronological order and the latter is an offline algorithm specially designed to
maximize the time at the top in feeds sorted in reverse chronological order.
However, because \redqueen{} assumes that the feed is inverse chronologically sorted and posts with intensity $\propto \text{rank}_{\text{chrono}}(t)$, we also compare our method TPPRL against a stronger heuristic RQ$^*$, which posts with intensity $\propto \text{rank}_{\text{priority}}(t)$.
%

%  We will train our neural network on all windows of size $T$ which lie in the two months (training set), except the last window, where we will test the performance of our agent.
    %
%    We will construct a batch of size $b$ to train our neural network by taking $b$ random windows of size $T$ from the training set, replaying the tweets in the feed of the followers (the feedback events) while our agent
% posts tweets (acts), and calculating the reward from the final trajectory of the system.
    %
%     The reward function and the feed sorting mechanism will change for different settings below.

\xhdr{Results}
Figures~\ref{fig:results-feed-sorting-algorithm}(a-b) summarize the results, where the number of messages posted by each method is the same and all rewards are normalized by the reward achieved by a
baseline user who follows a uniform Poisson intensity. The results show that, by not making any assumption about the feed sorting algorithm, our method is able to outperform both \redqueen{} and Karimi'{}s method,
which were specially designed to minimize the average rank and time at the top in feeds sorted in reverse chronological order, respectively.
Moreover, our method provides solutions with smaller variance in performance than \redqueen{}.
Finally, in Figure~\ref{fig:results-feed-sorting-algorithm}(c), we give some intuition on the type of policy our method learns using a toy example, where a user competes for attention with two other users
in a follower'{}s feed, one with higher priority and another with lower priority. Our method learns to avoid posting whenever the user with higher priority posts.

\section{Conclusions}
\label{sec:conclusions}
% \vspace{-4mm}
In this paper, we approached a novel reinforcement learning problem where both actions and feedback are asynchronous
stochastic events in continuous time, characterized using marked temporal point processes (MTPPs).
In this problem, the policy is a conditional intensity function (and mark distribution), which is then used to sample the times
(and marks) of the agent'{}s actions.
Then, we derived a flexible policy gradient method, which does not make any assumptions on the functional form of the intensity
and mark distribution of the feedback and it allows for arbitrarily complex reward functions.
%
% Our method does not make any assumptions on the functional form of the intensity and mark distribution of the feedback and
% it allows for arbitrarily complex reward functions.
%
Experiments on two different applications in personalized teaching and viral marketing show that our method beats competing
methods.

There are many interesting venues for future work. For example, we have taken a first step towards developing reinforcement learning
algorithms for MTPPs, however, a natural follow up would be deriving more sophisticated reinforcement learning algorithms, \eg,
actor-critic algorithms, for our novel problem setting.
We have evaluated in two real-world applications in personalized teaching and viral marketing, however, there are many other (high
impact) applications fitting our novel problem setting, \eg, quantitative trading.
Finally, it would be very interesting to develop multiple agent reinforcement learning algorithms for MTPPs.
%
%
% \paragraph{Future work:}
% \begin{itemize}
%  \item Extending to actor-critic framework.
%  \item Including some modeling assumptions as part of the model.
% \end{itemize}

\clearpage
\newpage
% \section*{References}
{
% \vspace{-3mm}
% \small
\bibliographystyle{abbrv}
\bibliography{refs}
}

\newpage

\appendix
% manuel: I moved the likelihood to the main
% \section{Additional Background on Marked Temporal Point Processes}\label{sec:mtpp}
%
% Given a marked temporal point process $\Hcal$ with joint probability density $p^{*} = (\lambda^{*}, m^{*})$, the likelihood
%of a history of events $\history{T} = \{ e_i = (t_i, z_i) \in \Hcal \st t_i < t \}$ under the MTPP is given by~\cite{Aalen2008}:
%
%\begin{equation}
%\PP(\history{T}) = \left( \prod_{e_i\in \history{T}} \lambda^{*}(t_i) m^{*}(z_i) \right) \exp \left( \int_{0}^{T} \lambda^{*}(s)\,ds \right).
%\end{equation}
%
%where, for each event $e_i = (t_i, z_i)$ the sign $^{*}$ denotes an implicit (potential) dependence on the history $\Hcal_{t_i}$.
%
{\Large \bf Appendix}

\section{Proof of Proposition~\ref{prop:gradient}}\label{app:gradient}
We first start by rewriting the expected reward function $J(\theta)$ as:
\begin{align*}
J(\theta) &= \EE_{\historyAgent{T} \sim \policyTheta(\cdot), \historyFeedback{T} \sim \feedback(\cdot)} \left[ \reward(T) \right] = \EE_{|\historyAgent{T}|, |\historyFeedback{T}| } \left[ \EE_{\historyAgent{T}, \historyFeedback{T} \,|\, |\historyAgent{T}|, |\historyFeedback{T}|} \left[R(T) \,|\, |\historyAgent{T}|, |\historyFeedback{T}|\right]\right]  \\
&= \sum_{m, k} \prob(|\historyAgent{T}| = m) \left( \prod_{i \in \historyAgent{T}} \int_{t_i, y_i} \uStar(t_i) \vStar(y_i) \right) \exp \left( - \int_{0}^{T} \uStar(s)\,ds \right) \\
& \times \prob(|\historyFeedback{T}| = k) \left( \prod_{j \in \historyFeedback{T}} \int_{t_j, z_j} \uFeedback(t_j) \vFeedback(z_j) \right) \exp \left( - \int_{0}^{T} \uFeedback(s)\,ds \right) \reward(T) \prod_{i \in \historyAgent{T}} d\,t_i d\,y_i  \prod_{j \in \historyFeedback{T}} d\,t_j d\, z_j,
\end{align*}
where we have first taken the expectation with respect to all histories conditioned on a given number of events and then taken the expectation with respect to the number
of events.
Then, we can compute the gradient $\grad_{\theta} J(\theta)$ as follows:
\begin{align*}
\grad_{\theta} J(\theta) &= \sum_{m, k} \grad_{\theta} \left\{ \prob(|\historyAgent{T}| = m) \left( \prod_{i \in \historyAgent{T}} \int_{t_i, y_i} \uStar(t_i) \vStar(y_i) \right) \exp \left( - \int_{0}^{T} \uStar(s)\,ds \right) \right\} \\
& \times \prob(|\historyFeedback{T}| = k) \left( \prod_{j \in \historyFeedback{T}} \int_{t_j, z_j} \uFeedback(t_j) \vFeedback(z_j) \right) \exp \left( - \int_{0}^{T} \uFeedback(s)\,ds \right) \reward(T) \prod_{i \in \historyAgent{T}} d\,t_i d\,y_i  \prod_{j \in \historyFeedback{T}} d\,t_j d\, z_j \\
&= \sum_{m, k} \frac{\grad_{\theta} \left\{ \prob(|\historyAgent{T}| = m) \left( \prod_{i \in \historyAgent{T}} \int_{t_i, y_i} \uStar(t_i) \vStar(y_i) \right) \exp \left( - \int_{0}^{T} \uStar(s)\,ds \right) \right\}}{\prob(|\historyAgent{T}| = m) \left( \prod_{i \in \historyAgent{T}} \int_{t_i, y_i} \uStar(t_i) \vStar(y_i) \right) \exp \left( - \int_{0}^{T} \uStar(s)\,ds \right)} \\
& \times \prob(|\historyAgent{T}| = m) \left( \prod_{i \in \historyAgent{T}} \int_{t_i, y_i} \uStar(t_i) \vStar(y_i) \right) \exp \left( - \int_{0}^{T} \uStar(s)\,ds \right) \\
& \times \prob(|\historyFeedback{T}| = k) \left( \prod_{j \in \historyFeedback{T}} \int_{t_j, z_j} \uFeedback(t_j) \vFeedback(z_j) \right) \exp \left(- \int_{0}^{T} \uFeedback(s)\,ds \right) \reward(T) \\
& \times \prod_{i \in \historyAgent{T}} d\,t_i d\,y_i  \prod_{j \in \historyFeedback{T}} d\,t_j d\, z_j \\
&= \sum_{m, k} \grad_{\theta} \left\{ \log \left( \prob(|\historyAgent{T}| = m) \left( \prod_{i \in \historyAgent{T}} \int_{t_i, y_i} \uStar(t_i) \vStar(y_i) \right) \exp \left( - \int_{0}^{T} \uStar(s)\,ds \right) \right) \right\} \\
& \times \prob(|\historyAgent{T}| = m) \left( \prod_{i \in \historyAgent{T}} \int_{t_i, y_i} \uStar(t_i) \vStar(y_i) \right) \exp \left( - \int_{0}^{T} \uStar(s)\,ds \right) \\
& \times \prob(|\historyFeedback{T}| = k) \left( \prod_{j \in \historyFeedback{T}} \int_{t_j, z_j} \uFeedback(t_j) \vFeedback(z_j) \right) \exp \left(- \int_{0}^{T} \uFeedback(s)\,ds \right) \reward(T) \\
& \times \prod_{i \in \historyAgent{T}} d\,t_i d\,y_i  \prod_{j \in \historyFeedback{T}} d\,t_j d\, z_j \\
&= \EE_{\historyAgent{T} \sim \policyTheta(\cdot), \historyFeedback{T} \sim \feedback(\cdot)} \left[ \reward(T) \grad_{\theta} \log{\PP_{\theta}(\historyAgent{T})} \right]
\end{align*}
where we have used that $\frac{\grad_{\theta} f(\theta)}{f(\theta)} = \grad_{\theta} \log f(\theta)$ and
$$\log{\PP_{\theta}(\historyAgent{T})} = \sum_{e_i\in \actions{T}}\left(\log{\uStar(t_i)} + \log{\vStar(z_i)} \right) - \int_{0}^{T}\uStar(s)\,ds.$$

\section{Proof of Proposition~\ref{prop:gradient-regularized}}\label{app:gradient-regularized}
We first start by rewriting the penalized expected reward function $J_r(\theta)$ as:
\begin{align*}
J_r(\theta) &= \EE_{\historyAgent{T} \sim \policyTheta(\cdot), \historyFeedback{T} \sim \feedback(\cdot)} \left[ \reward(T) - q_l \int_0 ^T g_{\lambda}(\uStar(t))dt - q_m \int_0 ^T g_{m}(\vStar(t))dt\right] \\
& = \EE_{\historyAgent{T} \sim \policyTheta(\cdot), \historyFeedback{T} \sim \feedback(\cdot)} \left[ \reward(T) \right] - q_l \, \EE_{\historyAgent{T} \sim \policyTheta(\cdot), \historyFeedback{T} \sim \feedback(\cdot)} \left[\int_0 ^T g_\lambda(\uStar(t))dt\right] \\
& - q_m \, \EE_{\historyAgent{T} \sim \policyTheta(\cdot), \historyFeedback{T} \sim \feedback(\cdot)} \left[\int_0 ^T g_m(\vStar(t))dt\right],
\end{align*}
where we have just used the linearity of the expectation. Then, we can use Proposition~\ref{prop:gradient} and the chain rule to compute the gradient
$\grad_{\theta} J_r(\theta)$:
\begin{align*}
\grad_{\theta} J_r(\theta) &= \EE_{\historyAgent{T} \sim \policyTheta(\cdot), \historyFeedback{T} \sim \feedback(\cdot)} \left[ \reward(T)  \grad_{\theta} \log{\PP_{\theta}(\historyAgent{T})} \right] \\
&- q_l   \EE_{\historyAgent{T} \sim \policyTheta(\cdot), \historyFeedback{T} \sim \feedback(\cdot)} \left[\int_0 ^T g_\lambda(\uStar(t)) dt \, \grad_{\theta} \log{\PP_{\theta}(\historyAgent{T})} \right] \\
&- q_l   \EE_{\historyAgent{T} \sim \policyTheta(\cdot), \historyFeedback{T} \sim \feedback(\cdot)} \left[\int_0 ^T g'_\lambda(\uStar(t)) \grad_{\theta} \uStar(t) dt \right] \\
&- q_m   \EE_{\historyAgent{T} \sim \policyTheta(\cdot), \historyFeedback{T} \sim \feedback(\cdot)} \left[\int_0 ^T g_m(\vStar(t))dt \, \grad_{\theta} \log{\PP_{\theta}(\historyAgent{T})} \right] \\
&- q_m   \EE_{\historyAgent{T} \sim \policyTheta(\cdot), \historyFeedback{T} \sim \feedback(\cdot)} \left[\int_0 ^T g'_m(\vStar(t)) \grad_{\theta} \vStar(t) dt \right] \\
&= \EE_{\historyAgent{T} \sim \policyTheta(\cdot), \historyFeedback{T} \sim \feedback(\cdot)} \big[ \\
& \qquad\qquad\quad \left( \reward(T) - q_l \int_{0}^{T} g_\lambda(\uStar(t))\,dt - q_m \int_{0}^{T} g_m(\vStar(t))\,dt \right) \grad_{\theta} \log{\PP_{\theta}(\historyAgent{T})} \nonumber\\
  & \qquad\qquad - \left. \left(  q_l \int_{0}^{T} g'_\lambda(\uStar(t)) \grad_{\theta} \uStar(t)\,dt + q_m \int_{0}^{T} g'_m(\vStar(t)) \grad_{\theta} \vStar(t)\,dt \right) \right]
\end{align*}
where $g'_\lambda(\uStar(t)) = \frac{d\, g_\lambda(\uStar(t))}{d\, \uStar(t)}$ and $g'_m(\vStar(t)) = \frac{d\, g_m(\vStar(t))}{d\, \vStar(t)}$.

\section{Sampling event times from the intensity $\uStar(t)$} \label{app:sampling}

Immediately after taking an action at time $t_i$, the agent has to determine the time of the next action $t_{i+1}$ by sampling from the intensity function $\uStar(t)$ given
by Eq.~\ref{eq:alt-intensity}.
%
% , to determine the time of the next action $t_{i+1}$.
%
However, if a feedback event arrives at time $s < t_{i+1}$, \ie, the feedback event arrives \emph{before} the agent has performed her next action, then the intensity function
$\uStar(t)$ will need to be updated and the time $t_{i+1}$ will not be a valid sample from the updated intensity.
To overcome this difficulty, we design the following procedure, which to the best of our knowledge, is novel in the context of temporal point processes.
Recall that the intensity function of the action events was
\begin{align}
 \uStar(t)=\exp(b_\lambda+\Vb_h \hb_i)\exp(\omega_t (t-t_i))
\end{align}
In other words, we write $\uStar(t)=c.e^{\omega_t (t-t_i)}$ and $c$ changes due to an arrival of an event.
So, we can state our problem as the following more general problem of sampling from a partially known intensity function:
\begin{equation}
  \lambda(t) = \begin{cases}
    c_1 e^{-\omega (t - t_i)} & \text{ if } t < s\\
    c_2 e^{-\omega (t - t_i)} & \text{ otherwise, }\\
  \end{cases}
  \label{eqn:sampling-intensity}
\end{equation}
where the parameters $c_1$ is known to us at time $t_i$ but $s, c_2$ are revealed to us only at time $s$, \ie, if our sampled time is
greater than $s$.
Due to this, we cannot sample from the above intensity using simple rejection sampling or the superposition property of poisson processes, as previous
work~\cite{tabibian2017optimizing, redqueen17wsdm}.
Instead, at a high level, we solve the problem by first sampling a uniform random variable $u \sim U[0, 1]$ and then using it to calculate ${t}_{i+1} = CDF_1^{-1}(u \given c_1, t_i)$,
where $CDF_1(t \given c_1, t_i)$ denotes the cumulative distribution function of the next event time.
Here, we are using inverse transform sampling under the assumption that the intensity function is defined completely using $c_1$ only.
Then, we wait until the earlier of either ${t}_{i+1}$, when we \emph{accept} the sample, or $s$, in which case the parameters $c_2$ are
revealed to us.
With the full knowledge of the intensity function, we can now \emph{refine} our sample ${t}_{i + 1} \leftarrow CDF_2^{-1}(t \given c_1, t_i, c_2, s)$ re-using the
same $u$ that we had originally sampled.

To be able to perform the above procedure in an efficient manner, we should be able to express $CDF_1^{-1}(t \given c_1, t_i)$ and $CDF_2^{-1}(t \given c_1, t_i, c_2, s)$
analytically.
Perhaps surprisingly, we can indeed express both functions analytically for our parametrized intensity function, given by Eq.~\ref{eqn:sampling-intensity}, \ie,
\begin{align}
  CDF_1(t \given c_1, t_i) &= \Pr\left[\text{An event happens before } t\right] \nonumber \\
  &= 1 - \Pr\left[\text{No event in } (t_i, t] \right] \nonumber \\
  &= 1 - \exp{\left( -\int_{t_i}^t \lambda(\tau) d\tau \right) } \nonumber \\
  &= 1 - \exp{\left( -\int_{t_i}^t c_1 e^{-\omega(\tau - t_i) }d\tau \right) } \nonumber \\
  &= 1 - \exp{\left( \frac{c_1}{\omega} (e^{-\omega(t - t_i)} - 1) \right)} \nonumber \\
  \implies CDF_1^{-1}(u \given c_1, t_i) &= t_i - \frac{1}{\omega}\log{\left( 1 + \frac{\omega}{c_1}\log{(1 - u)} \right)} \label{eqn:cdf-1}\\
  \text{Similarly, } CDF_2^{-1}(u \given c_1, t_i, c_2, s) &= s - \frac{1}{\omega}\log{\left( 1 + \frac{\omega}{c_2}\log{\left( \frac{1 - u}{Q} \right)} \right)} \label{eqn:cdf-2}\\
  \text{where } Q &= \exp{\left( -\frac{c_1}{\omega}\left( 1 - \exp{\left( -\omega (s - t_i) \right)} \right) \right)}. \nonumber
  % \label{eqn:cdfs}
\end{align}
Notice that Eq.~\ref{eqn:cdf-2} is the same as Eq.~\ref{eqn:cdf-1}, if our uniform sample had been $u' = 1 - \frac{1 - u}{Q}$, and we had started the sampling process at time $s$ instead of time $t_i$ with parameters $c_2, \omega$.
Using this insight, we can easily generalize this sampling mechanism to account for an arbitrary number of feedback events occurring between two actions of the agent.
Algorithm~\ref{alg:sampling} summarizes our sampling algorithm, where \textsc{ComputeC1} and \textsc{ComputeC2} compute the current values of $c_1$ and $c_2$, respectively,
\textsc{WaitUntilNextFeedback}($t$) sets a flag $e$ to True if a feedback event $(s,z)$ happens before time $t$.
Remarkably, given a cut-off time $T$, the algorithm only needs to sample $|\Acal_T|$ times from a uniform distribution and perform $O(|\Hcal_T|)$ computations.
%
% Furthermore $ {CDF}^{-1}(u \given Q,c,t)$ unifies $CDF_2 ^{-1}$ and $CDF_1 ^{-1}$ (when $Q=1$).

Finally, note that, in the above procedure, there is a possibility that the inverse CDF functions may not be completely defined on the domain $[0, 1]$. This would mean that the agent'{}s MTPP may go \emph{extinct}, \ie,
there may be a finite probability of the agent not taking an action after time $t_i$ at all.
In such cases, we assume that the next action time is beyond our episode horizon $T$, but we will save the original $u$ and will keep calculating the inverse CDF using it as, due to the non-linear dependence of
the parameters on the history, the samples may become finite again.

\begin{algorithm}[t]                    % enter the algorithm environment
	\small
	\caption{It returns the next action time}
	\label{alg:sampling}
	\begin{algorithmic}[1]
	\STATE \textbf{Input: } Time of previous action $t'{}$, history $\Hcal_{t'{}}$ up to $t'{}$, cut-off time $T$
	\STATE \textbf{Output: } Next action time $t$

%	\STATE\texttt{/* Takes $\Hcal_{t_\text{last}}$ and final time $T$ as input, and samples the next action. */}
% 		\STATE $t_{\text{last}}\leftarrow 0$
% 		\WHILE{$t_{\text{last}}<T$}
		 \STATE $c_1 \leftarrow \textsc{ComputeC1}(\Hcal_{t'{}})$
		 \STATE $t \leftarrow {CDF}_{1}^{-1}(u \given c_1,t'{})$
% 		 \STATE $s\leftarrow t_{\text{last}}$
		  \WHILE{$t < T$}
		  	\STATE $(e,s,z)\leftarrow$\textsc{WaitUntilNextFeedback}$(t)$
		  	\IF{$e==\text{True}$}
		  		% \STATE $t'{} \leftarrow s$
		  		\STATE $\Hcal_{t'{}}\leftarrow \Hcal_{t'{}} \cup \{ (s,z) \}$
				\STATE $c_1\leftarrow\textsc{ComputeC1}(\Hcal_{t'{}})$, $c_2\leftarrow\textsc{ComputeC2}(\Hcal_{t'{}})$
				\STATE $t \leftarrow CDF_2^{-1}(u \given c_1, t'{}, c_2, s)$
		  	\ELSE
		  		% \STATE \textsc{TakeAction}$(t)$
				\RETURN t
% 		  \STATE $t_{\text{last}} \leftarrow t$
% 		  \STATE $Q\leftarrow 1$
  		  		\STATE \textbf{break}
		   \ENDIF
		  \ENDWHILE
	  \RETURN t
% 		  \ENDWHILE
	\end{algorithmic}
% 	\label{alg:opDynamicsSim}
\end{algorithm}
\begin{table*}[t]
 		\vspace{2mm}
		\small
		\centering
		 \scalebox{0.8}{
		\begin{tabular}{l|l|l|l|l|l|l|l|l}
		\hline
				{Application} &  {$N_b$}&  $N_e$ & $T$ & $l_r$ & $D_i$ & $D_h$ & $q_l$ & $q_m$ \\
				\hline\hline
				% {LL} & {-1.631} & -1.658 & -1.980 & -2.153 \\
 				{Spaced repetition} &{$5000$}&  $32$ & $14$ days & $\frac{0.02}{1 + 2i \cdot 10^{-3}}$ & $8$ & $8$ & $10^{-2}$ & $5 \cdot 10^{-3}$ \\
 				{Smart broadcasting} &{$1000$}&  $16$ & It varies across users & $\frac{10^{-2}}{1 + i \cdot 10^{-4}}$ & $8$ & $8$ & $0.33\ (100)$ & --\\

 				\hline
		\end{tabular}
	}
		%\caption{Results finding best molecule. Column 4 indicates fraction of sampled molecules with score greater than 0 }
		%\end{tabular}}
		\caption{Hyperparameter values used in the implementation of our method for smart broadcasting and spaced repetition. In smart broadcasting, $q_l=0.33$ for top-1 inverse chronological ordering
		and $q_l=100$ for average rank inverse chronological ordering.} % using our model, JTVAE ~\cite{jin2018junction}, GrammarVAE~\cite{kusner2017grammar} and CVAE~\cite{gomez2016automatic}.}
		% , GraphVAE~\cite{graphvae} does not report anythigs VVAE~\cite{janz2017learning} does not do BO
		% \ourmodel beats the state of the art by large margins except JTVAE ~\cite{jin2018junction}.}
		\label{tab:hyper}
\end{table*}
% We will solve this Hence, we ought to draw a sample from the process at time $t_i$ and optimistically
%
% However, we do not \emph{know} where
%
% To do this, we can invert the cumulative distribution function of the next arrival time
%
% We can do that by drawing a uniform random number $u \sim Unif[0, 1]$ and then inverting the
%
% It is easy to show that the intensity function in Eqs.~\ref{eq:alt-intensity} can be cast as the following more general problem form where the parameters $c_1, \omega_1$ can be derived given the current history of the process.

%
% To recall, the intensity function $\uStar(t)$ is given by:
% \begin{equation}
%   \uStar(t) =  \exp{\left( b_{\lambda} + w_t (t - t_i) + \Vb_{\lambda} \hb_i \right)}\nonumber
% \end{equation}
%
% From the point of view of the agent, the
%
% We produce samples with intensity $u_{\theta}(t)$ by inverting the CDF. \red{TODO by Utkarsh}

% \red{Sections below TODO by Utkarsh}

% \section{Viral marketing experimental details} \label{app:implementation}
% %
\section{Experimental details} \label{app:implementation}
\begin{figure}[t]
  \centering
  \subfloat[$J(\theta)$ for quadratic loss]{\includegraphics[width=0.4\textwidth]{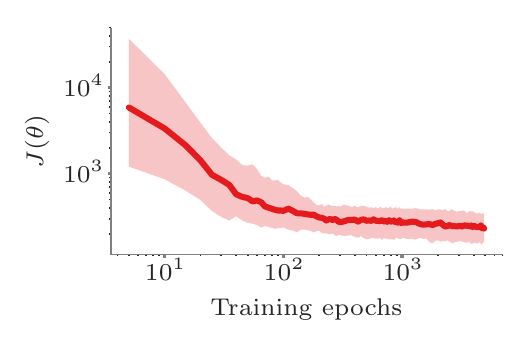}}\hspace{0.05\textwidth}%
  \subfloat[$J(\theta)$ for time spent at top]{\includegraphics[width=0.4\textwidth]{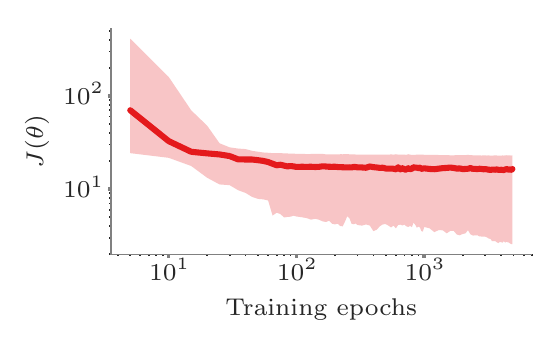}}
  \caption{The cost-to-go $J(\theta)$ calculated on the held-out test-set for different loss functions during training falls quickly with the number of epochs.}
  \label{fig:learning-error}
\end{figure}

We carried out all our experiments using TensorFlow 1.8.1 and we implemented stochastic gradient descent (SGD) using the Adam optimizer, which achieved good performance in practice, as shown in Figure~\ref{fig:learning-error}.
%
% Specifically, for each broadcaster, we split the feedback events she receives, into training and test sets-- where the test set consists of roughly 300 events.
%
% Specifically, we group the training events into $N$ batches and split these batches in  $e$  {episodes} of length $T$.
% %
% Thereafter, for each batch, we compute the mean of  all gradients $\grad_\theta J(\theta)$  over the underlying episodes.
% we grouped the events in training-data in $1000$ batches where each batch has
Therein, we had to specify eight hyperparameters:
(i) $N_b$ -- the number of batches,
(ii) $N_e$ -- the number of episodes in each batch,
(iii) $T$ -- the time length of each episode,
(iv) $l_r$-- the learning rate,
(v) $D_i$  -- the dimension of vectors $\Wb_\bullet, \bb_\bullet$'s in the input layer,
(vi) $D_h$  -- the dimension of the hidden state $\hb_i$,
(vii) $q_l$ -- the value of the regularizer coefficient for intensity function,
(viii) $q_m$ -- the value of the regularizer coefficient for mark distribution.
Note that, the dimensions of the other trainable parameters $\Wb_{h}, \Wb_1,..,\Wb_4$ and $\bb_h$ in the hidden layer depend on $D_i$ and $\Vb_{\lambda}$ and $\Vb^{y}_c$ in the output layer depend on $D_h$,
which we selected using cross validation. The values for both applications---spaced repetition and smart broadcasting ---are given in Table~\ref{tab:hyper}.

We run the spaced repetition experiments using a Tesla K80 GPU on a machine with 32 cores and 500GB RAM. With this configuration, for episodes with up to $\sim$$2000$ events, the training process
takes $\sim$$5$ seconds in average to run one iteration of SGD with batch size $N_e = 32$.
We run the smart broadcasting experiments on 2 CPU cores of an Intel(R) Xeon(R) CPU E5-2680 v2 @ 2.80GHz and 20GB RAM. With this configuration, for feeds sorted algorithmically and episodes with up
to $\sim$$250$ events, the training process takes $\sim$$30$ seconds to run one iteration of SGD with batch size $N_e = 16$.

% In the following, we present the values of these hyperparameters used for the two applications  considered in this paper.
%
% %
% \xhdr{Smart broadcasting}
% (i) $N_b=1000$,
% (ii) $N_e=16$,
% (iii) $T$ varies across different  broadcasters,
% (iv) $l_r=\frac{10^{-2}}{1 + i \times 10^{-4}}$,
% %
% (v) $D_i=8$,
% %
% (vi) $D_h=8$,
% %
% (vii) $q_l=0.33\ (100)$ for  top-1 inverse chronological ordering/algorithmic feeds (average rank inverse chronological ordering).
% %
% % The resulting values were $D_i=8$, $D_h=8$ and $q=0.33\ (100)$ for  top-1 inverse chronological ordering/algorithmic feeds (average rank inverse chronological ordering).
% Figure~\ref{fig:learning-error} shows  that $J(\theta)$ decreases smoothly with number of training epochs, for both reward functions.
%
% \xhdr{Memorize} (i) $N_b=5000$,
% (ii) $N_e=32$,
% (iii) $T=14$ days,
% (iv) $l_r=\frac{0.02}{1 + 2i\times 10^{-3}}$,
% %
% (v) $D_i=8$,
% %
% (vi) $D_h=8$,
% %
% (vii) $q_l=0.0001$,
% (viii) $q_m=0.005$
%

\section{Student model}\label{app:student-model}
We use the student model proposed by Tabibian~\ea{}~\cite{tabibian2017optimizing}, which is an improved version of the student model proposed by Settles~\ea{}~\cite{settles2016trainable}.
To accurately predict the student'{}s ability to recall an item, the model accounts for the item difficulty, the history of reviews (and recalls) by the student, and the time since the last review.

% aim is to learn a set of $N$ items, each of which have
%
More formally, the probability $m_i(t)$ that an item $i$, which was last reviewed at time $\eta$, will be successfully recalled at time $t$ is given by:
\begin{equation}
  m_i(t) = e^{-n_i(t) \times (t - \eta)}
  \label{eqn:prob-recall}
\end{equation}
where $n_i(t)$ denotes the forgetting rate for the item $i$.
The rate of forgetting an item depends on the inherent difficulty of the item, denoted by $n_i(0)$, but also on whether the user was able to recall the item successfully in the past or not.
%
% The difficulty of an item $i$ is described using two parameters $\alpha_i, \beta_i$, which describe by how much the forgetting rate ought to change if the student reviews it correctly or incorrectly, \ie, if a review is scheduled at time $t$:
%
More specifically, the model has two additional parameters $\alpha$ and $\beta$, which determine by how much the forgetting rate ought to change if the student recall, or fails to recall, the
item on a review at time $t$, \ie,
\begin{equation}
  n_i(t) = \begin{cases}
    (1 - \alpha) \times n_i(t^-) & \text{ if recalled}\\
    (1 + \beta) \times n_i(t^-) & \text{ if forgotten}
  \end{cases}
  \label{eqn:forgetting-rate-update}
\end{equation}
In our work, the parameters $\alpha$ and $\beta$, as well as the initial item difficulty $n_i(0)$, are learned using historical learning data from Duolingo as in Tabibian~\ea{}~\cite{tabibian2017optimizing}.

Note that we have picked this student model for its simplicity but relatively good predictive power, as shown by previous work. Several other student models have also been proposed in literature, ranging
from exponential~\cite{ebbinghaus1885memory} to more recent multi-scale context models (MCM)~\cite{pashler2009predicting}, which are biologically inspired and can explain a wider variety of learning
phenomenon.
Since our methodology is agnostic to the choice of student model, it would be very interesting to experiment with other student models.

\section{Feed sorting algorithm} \label{app:feed-sorting-algorithm}
We use a feed sorting algorithm inspired by the \emph{in-case-you-missed-it} feature, which is now prevalent in a variety of social media sites, notably Twitter at
the time of writing.
%
% , our sorting algorithm assigns a \emph{priority} to each user the user follows. Then, it divides the feed
Our sorting algorithm divides each user'{}s feed in two sections:
(i) a prioritized section at the top of the user'{}s feed, where messages are sorted according to the \emph{priority} of the user who posted the message,
and (ii) a bulk section, where messages are sorted in reverse chronological order.
In the above, each post stays for a fixed time $\tau$ in the prioritized section and then it moves to the inverse chronological section. Moreover, note that if the prioritized
section contains several messages from the same user, they are sorted chronologically.
%
% This simple sorting algorithm allows us to move between purse reverse chronological order to purely prioritized feed by tuning the parameter $\Tcal$.
%
% \manuel{the paragraph below is redundant given (i-ii)
% , at a position which depends on the priority of the user who posted the message as well as on
% how many other (higher) priority users have messages which are less than $\Tcal$ time old (higher priority users stay on top).
%
% After a time $\Tcal$, the post moves back to the inverse chronological section, where the rank is determined by the recency of the post.
%smart broadcasting and
% This simple model of feed sorting manages to capture the major departures from inverse chronologically sorted feeds by modifying a single parameter $\Tcal$.

In our experiments, for each user'{}s feed, we set the priority of the users she follows inversely proportional to her level of activity, as more active users will naturally appear on the feed
while users with sporadic posting activity may need more promotion, we set the priority of the user under our control to be at the median priority among all users posting in the feed, and set $\tau$ to
be approximately $10$\% of
the prioritized lifetime of posts $\tau = 0.1 T$, where $T$ is the time length of each sequence.

\begin{figure}[t]
  \centering
  \subfloat[Average rank]{\includegraphics[width=0.3\textwidth]{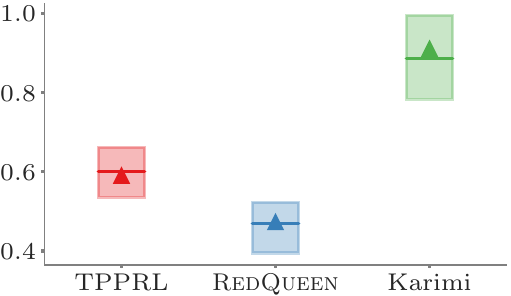}\label{fig:inv-avg-rank}}\hspace{0.04\textwidth}%
  \subfloat[Time at top]{\includegraphics[width=0.3\textwidth]{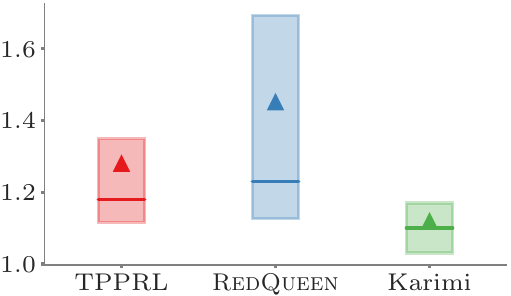}\label{fig:inv-top-1}}\hspace{0.04\textwidth}%
%  \subfloat[Example trace]{\includegraphics[width=0.3\textwidth]{example-755}\label{fig:trace}}
  \caption{Performance of our policy gradient method against \redqueen{}~\cite{redqueen17wsdm} and Karimi's method~\cite{karimi2016smart} on feeds sorted in reverse
  chronological order.
  Panels (a) and (b) show the average rank and time at the top, where the solid horizontal line shows the median value across users, normalized
  with respect to the value achieved by a user who follows a uniform Poisson intensity, and the box limits correspond to the $25$\%-$75$\% percentiles.
  For the average rank, lower is better and, for time at the top, higher is better. In both cases, the number of messages posted by each method is the
  same.
  %
%  Panel (c) shows a user'{}s intensity $\uStar(\cdot)$ (in blue), as provided by our method, the counts of the user'{}s posts (in green), and the average
%  rank (in red).
  }
  \label{fig:results-reverse-chronological-order}
\end{figure}

\section{Experiments on feeds sorted in reverse chronological order} \label{app:feed-reverse-chronological-order}
We follow the same experimental setup as in Section~\ref{sec:smart-broadcasting}, however, feeds are sorted in reverse chronological order.
Figure~\ref{fig:results-reverse-chronological-order} summarizes the results, where the number of messages posted by each method is the same
and all rewards  are normalized by the reward achieved by a baseline user who follows a uniform Poisson intensity.
The results show that our method is able to achieve competitive results in comparison with \redqueen{}, which is an online algorithm specially
designed to minimize the average rank in feeds sorted in reverse chronological order, and it outperforms Karimi'{}s method, which is an offline
algorithm specially designed to maximize the time at the top in feeds sorted in reverse chronological order.

% We describe the detailed experimental setup, including number of iterations, hyper-parameters, and the details of the mechanics of the algorithmic feed here. \red{TODO by Utkarsh}
%
% TODO:
% \begin{itemize}
%  \item Move to the integral notation.
%  \item Mention that we use the mean adjusted reward $\bar{\reward}(T, h) = \reward(T, h) - \frac{1}{n}\sum_{h \in \historyspace{T}} \reward(T, h)$ to reduce the variance of the gradients.
% \end{itemize}
%
%
% However, we will model the student using the somewhat simpler exponential model of memory~\cite{settles2016trainable}.
%
% \red{TODO}.

\section{Baseline with $w_t = 0$}

We also explored how our algorithm performs when we force the $w_t$ parameter to be zero, \ie, we force the policy to be piece-wise constant between feedback and action events.
To this end, we retrained the neural networks by doing a parameter sweep over $q_l$ (and $q_m$ for the spaced repetition experiments) and picked those values which arrived to roughly the same number of events as produced by the policy learned by the network where we do not constraint $w_t = 0$.

\begin{figure}[t]
  \centering
  \subfloat[Average rank]{\includegraphics[width=0.24\textwidth]{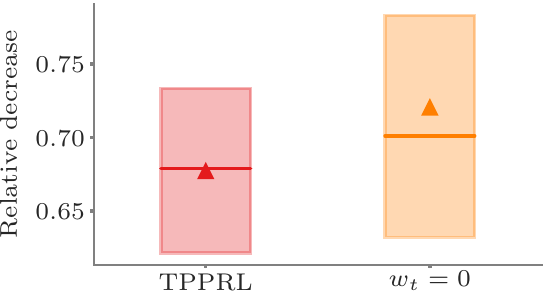}\label{fig:avg-rank-wt_zero}}\hspace{0.01\textwidth}%
  \subfloat[Time at top]{\includegraphics[width=0.24\textwidth]{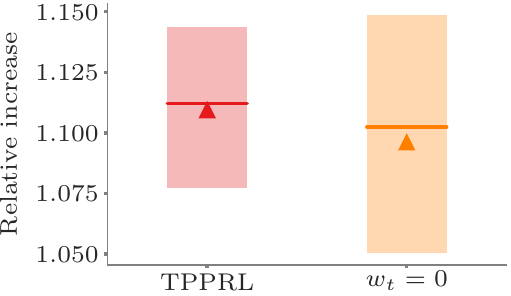}\label{fig:top-1-wt_zero}}\hspace{0.01\textwidth}%
  \subfloat[Recall]{\includegraphics[width=0.24\textwidth]{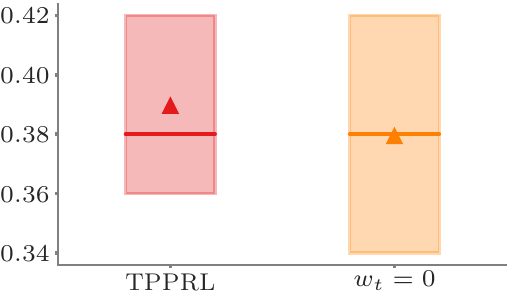}\label{fig:recall-wt_zero}}\hspace{0.01\textwidth}%
  \subfloat[Items' difficulty]{\includegraphics[width=0.24\textwidth]{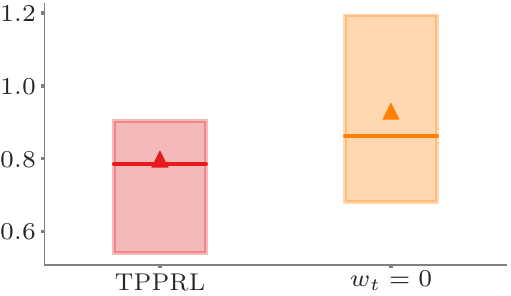}\label{fig:item-difficulty-wt_zero}}
  %
  % \subfloat[Example trace]{\includegraphics[width=0.3\textwidth]{example-755}\label{fig:trace}}
  \caption{Comparing against piece-wise constant ($w_t = 0$) baseline.
  In all figures, the solid horizontal line shows the median value across users and the box limits correspond to the $25$\%-$75$\% percentiles.
  Panels (a) and (b) show the average rank and time at the top for the smart broadcasting experiments, respectively.
  The values are normalized with respect to the value achieved by a user who follows a uniform Poisson intensity.
  For the average rank, lower is better and, for time at the top, higher is better.
  In both cases, the number of messages posted by each method is the same within a $10\%$ tolerance.
  Panel (c) shows the empirical recall probability at test time and Panel (d) shows the distribution of the difficulty of items chosen by our method and the baseline version for the space repetition experiments.
  The total number of learning events (across all items) are within $5\%$ of each other in the two settings.
}
  \label{fig:results-wt-zero}
\end{figure}

The resulting baseline is shown in Figure~\ref{fig:results-wt-zero} for both the smart broadcasting (Figures~\ref{fig:avg-rank-wt_zero} and \ref{fig:top-1-wt_zero}) and spaced repetition experiments (Figures~\ref{fig:recall-wt_zero} and \ref{fig:item-difficulty-wt_zero}).
We see that forcing the policy to be piecewise constant degrades performance and increases the variance in both settings, as expected.
In the smart broadcasting experiments, the mean (median) relative decrease in average rank is 33\% (33\%) for our method TPPRL, while it is 28\% (30\%) for the $w_t = 0$ baseline.
Similarly, the increase in mean time spent at the top is about 11\% for our method (TPPRL), while it is 9\% for the $w_t = 0$ baseline.
In the spaced repetition experiment, we see that the mean recall falls from 38.9\% to 37.9\%.
The difference in policy learned is especially notable in Figure~\ref{fig:item-difficulty-wt_zero} where we see that the agent, when constrained to $w_t = 0$, learns to spread its attempts over a wider set of items, which have higher difficulty than the items selected by the unconstrained policy.

\end{document}